\documentclass[11pt]{article}

\usepackage[final]{acl} 

\usepackage{times}
\usepackage{latexsym}

\usepackage{hyperref}
\usepackage{url}
\usepackage{wrapfig}
\usepackage{booktabs}
\usepackage{subcaption}
\usepackage{graphicx}
\usepackage{multirow}
\usepackage{float}
\usepackage{algorithm}
\usepackage{algorithmic}
\usepackage{amsmath,amsfonts,bm}

\usepackage{xcolor}
\usepackage{tikz}
\usetikzlibrary{arrows.meta,shadows,fit,calc,backgrounds}
\usepackage[edges]{forest}
\usepackage{graphicx}
\usepackage{ragged2e}

\usepackage[T1]{fontenc}

\usepackage[utf8]{inputenc}

\usepackage{microtype}

\usepackage{inconsolata}

\usepackage{graphicx}

\title{Efficient Task Adaptation in Large Language Models: A Survey of Weight-Based, Prompt-Based, and Embedding-Based Adaptations}

\author{
 \textbf{Jungwon Park$^{1,5}$},
 \textbf{Changin Choi$^{2,6}$},
 \textbf{Jimyeong Kim$^{3}$},
 \textbf{Nojun Kwak$^{2,3,4}$},
 \textbf{Wonjong Rhee$^{2,3,4}$\thanks{Corresponding Author}}
\\
 $^1$RICS, $^2$IPAI, $^3$AIIS, $^4$Department of Intelligence and Information, 
\\
 Seoul National University
\\ 
 $^5$Daegu Gyeongbuk Institute of Science and Technology 
\\
 $^6$Samsung Advanced Institute of Technology, Samsung Electronics Co., Ltd
\\
 \small{ \texttt{\{quoded97, ci2015.choi, wlaud1001, nojunk, wrhee\}@snu.ac.kr}}
}

\begin{document}
\addtocontents{toc}{\protect\setcounter{tocdepth}{-1}}
\maketitle
\begin{abstract} 
As large language models are increasingly deployed across diverse downstream tasks, efficient task adaptation has emerged as a central challenge. In response, a wide range of task adaptation methods have been proposed, spanning parameter-efficient fine-tuning, in-context learning, and embedding-injection approaches. However, these lines of work have largely evolved within individual paradigms, leaving their cross-paradigm relationships and trade-offs underexplored, especially for recently emerging embedding-based adaptations. This survey presents a unified framework that categorizes task adaptation methods by where and how task information is encoded: model weights, input prompts, or injected task embeddings. We provide a comprehensive taxonomy that integrates these paradigms, analyze their key strengths and limitations to explain how different adaptation paradigms have evolved, clarify relationships across paradigms, and highlight open problems for future research. \footnote{GitHub Repository: \url{https://github.com/Jungwona/Awesome-Task-Adaptation-for-Large-Language-Models}}
\end{abstract}
\section{Introduction}
\label{sec:introduction}
Large language models~(LLMs) are increasingly deployed across a wide range of downstream tasks~\citep{brown2020language, wei2022emergent, minaee2024large, raza2025industrial}, which makes efficient and effective task adaptation a central challenge for practical deployment. At the same time, as LLMs continue to scale in model size and training cost~\citep{kaplan2020scaling, hoffmann2022training, achiam2023gpt, openai2025, comanici2025gemini}, retraining or fully fine-tuning these models has become increasingly expensive and often infeasible, even when task-specific data are available. This growing gap between the demand for task adaptation and the cost of full fine-tuning has driven sustained interest in methods that reduce training cost, memory usage, and deployment overhead while maintaining strong task performance. Consequently, a wide range of approaches have been proposed, including methods that limit the number of trainable parameters or perform adaptation without any parameter updates. In this work, we focus on \emph{efficient task adaptation}, which we define as adapting pretrained or instruction-tuned LLMs to downstream tasks without full fine-tuning.

\definecolor{rootgreen}{RGB}{208,248,206}
\definecolor{methodtan}{RGB}{255,242,204}
\definecolor{leafwhite}{RGB}{255,255,255}
\definecolor{softteal}{RGB}{200,235,235}   
\definecolor{iceblue}{RGB}{225,240,248}    
\definecolor{softgold}{RGB}{245,235,200}   
\definecolor{lightmustard}{RGB}{242,230,185} 
\definecolor{sagegreen}{RGB}{215,230,205}  
\definecolor{mintgreen}{RGB}{205,240,225}  
\definecolor{warmgray}{RGB}{235,235,235}   
\definecolor{coolgray}{RGB}{225,230,235}   
\definecolor{softviolet}{RGB}{225,210,245} 
\definecolor{dustypurple}{RGB}{215,205,230} 
\definecolor{lavender}{RGB}{235,210,235}   
\definecolor{gray}{RGB}{220,220,220}   
\definecolor{palered}{RGB}{255,220,220}  
\definecolor{softcoral}{RGB}{255,210,200}  
\definecolor{softorange}{RGB}{255,225,200} 

\definecolor{coolgray}{RGB}{225,230,235}

\definecolor{softteal}{RGB}{200,235,235}
\definecolor{softtealA}{RGB}{220,245,245}  
\definecolor{softtealB}{RGB}{235,250,250}  

\definecolor{catblue}{RGB}{218,232,252}
\definecolor{catblueA}{RGB}{232,242,252}  
\definecolor{catblueB}{RGB}{242,248,254}  

\definecolor{lilacgray}{RGB}{232,228,240}  
\definecolor{lilacgrayA}{RGB}{242,240,248}
\definecolor{lilacgrayB}{RGB}{248,247,252}

\definecolor{softviolet}{RGB}{225,210,245}
\definecolor{softvioletA}{RGB}{235,225,250}
\definecolor{softvioletB}{RGB}{245,238,252}

\definecolor{softcoral}{RGB}{255,210,200}
\definecolor{softcoralA}{RGB}{255,225,218}
\definecolor{softcoralB}{RGB}{255,238,232}

\definecolor{sagegreen}{RGB}{215,230,205}
\definecolor{softgold}{RGB}{245,235,200}

\definecolor{embcol}{RGB}{210,230,255}   
\definecolor{parcol}{RGB}{255,230,190}   
\definecolor{procol}{RGB}{215,245,220}   

\begin{figure*}
\centering
\begin{forest}
  tikz={scale=0.68, transform shape},
  for tree={
    grow=east,
    forked edges,
    draw,
    rounded corners=3pt,
    align=center,
    edge={draw=black!60, -{Latex[length=2mm]}, line width=0.8pt},
    parent anchor=east,
    child anchor=west,
    anchor=west,
    drop shadow,
  },
  root style/.style={
    fill=coolgray,
    fit=band,
    font=\bfseries\small,
    inner sep=5pt,
    align=center,
  },
  levelone/.style={
    fill=catblue,
    fit=band,
    font=\bfseries\fontsize{8.1}{8.1}\selectfont,
    text width=3.2cm,
    inner sep=1pt,
    node options={align=center},
    minimum height=1.2cm,
  },
  leveltwo/.style={
    fill=methodtan,
    font=\bfseries\fontsize{7.2}{7.2}\selectfont,
    text width=4.5cm,
    node options={align=center},
    inner sep=2pt,
    minimum height=0.5cm,
  },
  levelthree/.style={
    fill=methodtan,
    font=\fontsize{6.9}{6.9}\selectfont,
    text width=4.0cm,
    node options={align=center},
    inner sep=2pt,
    minimum height=0.42cm,
  },
  longleveltwo/.style={
    fill=methodtan,
    font=\fontsize{6.2}{7.0}\selectfont,
    text width=\dimexpr\textwidth-4.37cm\relax,
    align=left,
    inner sep=2pt,
  },
  leaf/.style={
    fill=leafwhite,
    font=\scriptsize,
    align=left,
    inner sep=4pt,
  },
  emb branch/.style={
    fill=softteal,
    draw
  },
  emb subbranchA/.style={
    fill=softtealA,
    draw
  },
  emb subbranchB/.style={
    fill=softtealA,
    draw
  },
  par branch/.style={
    fill=catblue,
    draw
  },
  par subbranchA/.style={
    fill=catblueA,
    draw
  },
  par subbranchB/.style={
    fill=catblueB,
    draw
  },
  pro branch/.style={
    fill=lilacgray,
    draw
  },
  pro subbranchA/.style={
    fill=lilacgrayA,
    draw
  },
  pro subbranchB/.style={
    fill=lilacgrayB,
  },
  [{\rotatebox{90}{Efficient Task Adaptation in LLMs}}, root style
    [Sec.~\ref{sec:embedding-based adaptation} Embedding-Based\\\textbf{Adaptation}, levelone, emb branch
        [\textbf{Sec.~\ref{sec:learned task embeddings} Learned} \textbf{Task Embeddings}, leveltwo, emb subbranchA
            [\textbf{Other Learned Embeddings}, levelthree, emb subbranchB]
            [\textbf{Soft Prompting}, levelthree, emb subbranchB]
        ] 
        [\textbf{Sec.~\ref{sec:icl-derived task embeddings} ICL-Derived} \textbf{Task Embeddings}, leveltwo, emb subbranchA
            [\textbf{Contrastively Derived}, levelthree, emb subbranchB]
            [\textbf{Non-Contrastively Derived}, levelthree, emb subbranchB]
        ]
    ]
    [Sec.~\ref{sec:prompt-based adaptation} Prompt-Based\\Adaptation, levelone, pro branch
       [\textbf{Sec.~\ref{sec:demonstration optimization} Demonstration} \textbf{Optimization}, leveltwo, pro subbranchA
          [\textbf{Many-Shot ICL}, levelthree, pro subbranchB]
          [\textbf{Demonstration Generation}, levelthree, pro subbranchB]
          [\textbf{Selective Demonstration Annotation}, levelthree, pro subbranchB]
          [\textbf{Ordered Demonstration Selection}, levelthree, pro subbranchB]
          [\textbf{Unordered Demonstration Selection}, levelthree, pro subbranchB]
      ]
      [\textbf{Sec.~\ref{sec:instruction optimization} Instruction} \textbf{Optimization}, leveltwo, pro subbranchA
        [\textbf{Indirect Optimization}, levelthree, pro subbranchB]
        [\textbf{Direct Optimization}, levelthree, pro subbranchB]
      ]
    ]
    [Sec.~\ref{sec:weight-based adaptation} Weight-Based\\Adaptation, levelone, par branch
        [\textbf{Sec.~\ref{sec:multi-task adaptation} Multi-Task} \textbf{Adaptation}, leveltwo, par subbranchA
            [\textbf{Model Merging}, levelthree, par subbranchB]
            [\textbf{Multi-Task PEFT}, levelthree, par subbranchB]
        ]
        [\textbf{Sec.~\ref{sec:single-task adaptation} Single-Task} \textbf{Adaptation}, leveltwo, par subbranchA
            [\textbf{Other PEFT}, levelthree, par subbranchB]
            [\textbf{LoRA Variants}, levelthree, par subbranchB]
        ]
    ]
  ]
\end{forest}
\caption{\textbf{Overview of the Taxonomy of Efficient Task Adaptation in Large Language Models.} The main taxonomy branches are shown, while the full taxonomy with individual methods is provided in Figure~\ref{fig:taxonomy} of Appendix~\ref{sec:taxonomy_appendix}.}
\label{fig:taxonomy_reduced}
\end{figure*}
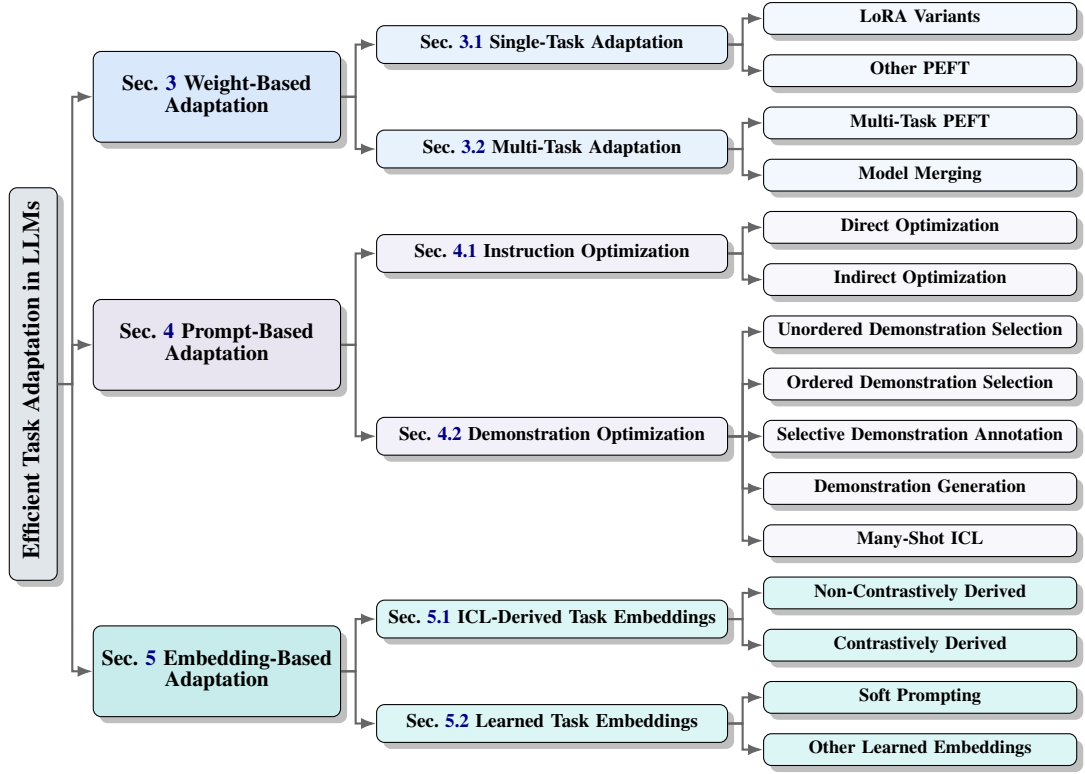

\begin{figure*}[t]
\begin{center}
    \includegraphics[width=\linewidth]{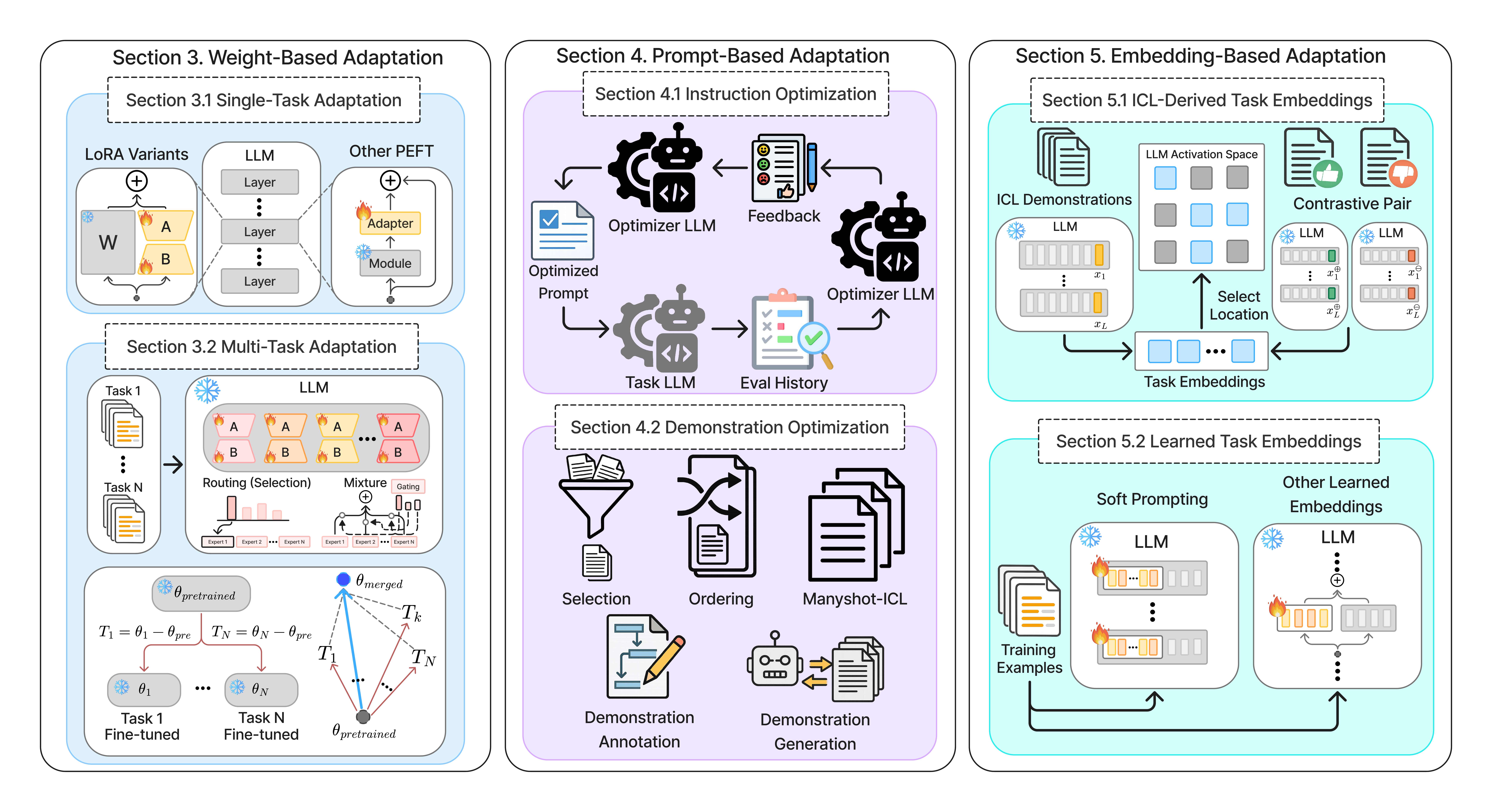}
\end{center}
\vspace{-5mm}
\caption{\textbf{Schematic Illustration of Three Task Adaptation Paradigms in LLMs.}} 
\label{fig:schematic illustration of three task adaptations}
\end{figure*}

Existing research has explored multiple paradigms for efficient task adaptation. Parameter-efficient fine-tuning~(PEFT)~\citep{houlsby2019parameter, lester-etal-2021-power, hu2022lora, liu2022fewshot} adapts models by training a small number of task-specific parameters while keeping the backbone model largely frozen. In-context learning~(ICL)~\citep{brown2020language, liu2022makes, lu2022fantastically, zhou2022large} enables training-free adaptation by specifying task information directly in prompts during inference. More recently, studies have shown that internal activations induced by ICL encode rich task-relevant information, giving rise to embedding-based adaptation that extracts and reuses explicit task representations during inference~\citep{hendel-etal-2023-context, todd2024function}. This line of work was originally developed in the context of mechanistic interpretability~\citep{olah2020zoom, elhage2021mathematical}, but has recently begun to be explored for practical deployment~\citep{li2025implicit, wang2025elicit, park2025softinjectiontaskembeddings}.

These approaches can be broadly categorized by where task information is encoded. We define three task adaptation paradigms from this perspective. \emph{Weight-based adaptation} encodes task information in model weights, including additional or modified parameters introduced by weight-based PEFT. \emph{Prompt-based adaptation} encodes task information in input prompts, such as task instructions or input-output demonstrations, without modifying the model. \emph{Embedding-based adaptation} encodes task information in explicit task embeddings that are injected into model activations during inference, where such embeddings can be derived from ICL or learned through optimization.

Despite rapid progress, research on task adaptation has largely evolved independently within each paradigm, leaving the relationships among weight-based, prompt-based, and embedding-based methods underexplored. While prior PEFT-only and ICL-only surveys provide useful taxonomies within their respective scopes~\citep{dong2024survey, sahoo2024systematic, vatsal2024survey, han2024parameter, wang2025parameter, mao2025survey}, they do not jointly organize these paradigms under a common task-adaptation perspective, especially for emerging ICL-derived embedding-based methods.

This separation makes it difficult to see how different adaptation strategies relate to one another and how their high-level trade-offs differ in terms of model access, training requirements, inference overhead, parameter efficiency, and deployment constraints. To address this gap, we present a unified survey that organizes task adaptation methods for LLMs according to where and how task information is encoded. Figure~\ref{fig:taxonomy_reduced} provides an overview of the main taxonomy branches, while the full taxonomy, including individual methods, is provided in Figure~\ref{fig:taxonomy} of Appendix~\ref{sec:taxonomy_appendix}. This survey makes the following contributions:
\vspace{-1.0mm}
\begin{itemize}
\item We propose a unified taxonomy of LLM task adaptation methods based on where and how task information is encoded, and analyze their connections, strengths, and limitations.
\vspace{-1.5mm}
\item We provide the first dedicated survey that includes recently emerging embedding-based adaptations and clarify their relationships to ICL and PEFT.
\vspace{-1.5mm}
\item We identify open challenges and future research directions for task adaptation in LLMs.
\end{itemize}

\section{A Unified View of Task Adaptation Paradigms}
\label{sec: A Unified View of Task Adaptation in Large Language Models}
\begin{table*}[t]
\centering
\resizebox{\textwidth}{!}{%
\begin{tabular}{l|l|l|l}
\toprule
\textbf{Adaptation Type} & \textbf{Task Information Encoding and Injection} & \textbf{Key Strengths} & \textbf{Key Limitations} \\
\toprule
\begin{tabular}[l]{@{}l@{}}
Weight-Based \\
(PEFT)
\end{tabular}
&
\begin{tabular}[c]{@{}l@{}}
\textbf{Loc:}~Task information encoded in model weights \\
\textbf{Build:}~Partial or additional weights trained \\
\textbf{Inject:}~Task-specific weights applied at inference
\end{tabular}
&
\begin{tabular}[c]{@{}l@{}}
Strong and stable task performance \\
Well established PEFT libraries \\
Lower inference overhead than ICL
\end{tabular}
&
\begin{tabular}[c]{@{}l@{}}
Requires training and parameter access \\
Task-specific modules must be loaded \\
Difficult mixed-task batching
\end{tabular}

\\
\midrule
\begin{tabular}[l]{@{}l@{}}
Prompt-Based \\
(ICL)
\end{tabular}
&
\begin{tabular}[c]{@{}l@{}}
\textbf{Loc:}~Task information encoded in prompts \\
\textbf{Build:}~Instructions or demonstrations in prompts \\
\textbf{Inject:}~Task-specific prompts used as input
\end{tabular}
&
\begin{tabular}[c]{@{}l@{}}
Training-free adaptation \\
Applicable to proprietary/API models \\
Natural-language task specification
\end{tabular}
&
\begin{tabular}[c]{@{}l@{}}
Repeated inference incurs optimization cost \\
Long prompts increase deployment cost \\
Sensitive to prompt formulation
\end{tabular}

\\
\midrule
\begin{tabular}[l]{@{}l@{}}
Embedding-Based \\
(ICL-derived)
\end{tabular}
&
\begin{tabular}[c]{@{}l@{}}
\textbf{Loc:}~Task information encoded in embeddings \\
\textbf{Build:}~Embeddings derived from demonstrations \\
\textbf{Inject:}~Embeddings injected into model activations
\end{tabular}
&
\begin{tabular}[c]{@{}l@{}}
Compact task representations \\
Lower inference overhead than ICL
\end{tabular}
&
\begin{tabular}[c]{@{}l@{}}
Requires access to internal activations \\
Requires task embedding construction
\end{tabular}

\\
\midrule
\begin{tabular}[l]{@{}l@{}}
Embedding-Based \\
(Learned)
\end{tabular}
&
\begin{tabular}[c]{@{}l@{}}
\textbf{Loc:}~Task information encoded in embeddings \\
\textbf{Build:}~Embeddings learned via optimization \\
\textbf{Inject:}~Embeddings injected into model activations
\end{tabular}
&
\begin{tabular}[c]{@{}l@{}}
Compact task representations \\
Lower inference overhead than ICL
\end{tabular}
&
\begin{tabular}[c]{@{}l@{}}
Requires access to internal activations \\
Optimization can be unstable \\
Training may require many iterations
\end{tabular}

\\
\bottomrule
\end{tabular}%
}
\vspace{-1.0mm}
\caption{\textbf{Conceptual Comparison of Task Adaptation Paradigms in LLMs.}
The table compares weight-based, prompt-based, and embedding-based adaptation methods. It describes where and how task information is encoded and injected into the model, followed by their key strengths and limitations, reflecting common characteristics observed in the literature. The column \emph{Task Information Encoding and Injection} specifies where task information is encoded~(Loc), how it is constructed~(Build), and how it is injected into the model during inference~(Inject).}
\vspace{-3.5mm}
\label{tab:adaptation_comparison}
\end{table*}

As defined in Section~\ref{sec:introduction}, we broadly categorize efficient task adaptation methods by where task information is encoded: model weights, input prompts, or explicit task embeddings injected into model activations. This perspective, illustrated in Figure~\ref{fig:schematic illustration of three task adaptations}, provides a useful lens for comparing the characteristics and goals of each paradigm.

Table~\ref{tab:adaptation_comparison} summarizes the key characteristics of these paradigms by describing where and how task information is encoded and injected into the model, followed by their key strengths and limitations. The table reflects representative trends rather than strict properties of individual methods and provides high-level insights into the design differences and trade-offs among adaptation strategies. The following paragraphs further explain how these strengths and limitations are derived from the literature.

\paragraph{Weight-Based Adaptation.} Prior work has shown that (weight-based) PEFT methods can achieve strong and stable performance~\cite{mahabadi2021compacter, mosbach2023few, weyssow2025exploring}, often comparable to full fine-tuning~\citep{hu2022lora, liu2022fewshot}. In addition, well-established PEFT libraries are widely available for various models, making these methods easily accessible to practitioners~\cite{peft}. However, these methods require access to model parameters and involve training additional or partial weights, which incurs memory and computation costs during training. They may also introduce adapter-management overhead in serving systems, since task-specific modules must be dynamically loaded and scheduled across requests~\cite{sheng2023s,chen2024punica}. In particular, mixed-task batching can be challenging when prompts within the same batch correspond to different tasks and require activating different task-specific modules. Serving such heterogeneous requests often requires specialized kernels, scheduling policies, and pipelined execution to mitigate latency~\cite{wu2024dlora,li2025toppings}.

\paragraph{Prompt-Based Adaptation.} Prompt-based adaptation interacts with the task-solving LLM only through its standard input-output interface and therefore does not require access to its parameters or internal activations. This makes it applicable to proprietary or API-only models such as GPT-5~\citep{openai2025} and GPT-6~\citep{openai2026}. Natural-language prompts are also human-readable and readily editable, allowing users to directly specify and modify task instructions or demonstrations. Beyond manual prompt design, prompts can be optimized in a black-box manner, with an optimizer LLM deriving feedback from the task LLM's evaluation history and using it to iteratively refine the prompt~\citep{zhou2022large, Pryzant2023AutomaticPO, yang2023large}. Recent work has extended this feedback-driven optimization framework beyond task instructions to agent skills, which encode reusable procedural knowledge and tool-use strategies as textual artifacts~\citep{alzubi2026evoskill, yang2026skillopt}. However, prompt-based adaptation can incur substantial costs from repeated LLM inference during prompt optimization and from the additional prefill computation and KV-cache memory required by long prompts at deployment~\cite{choi2023fixed, liskavets2025prompt, hooper2025squeezed}. Furthermore, because prompts serve as the primary means of task specification, performance can vary substantially with instruction formulation and demonstration selection or ordering, potentially requiring careful prompt construction or iterative prompt optimization to achieve reliable performance~\citep{zhou2022large, liu2022makes, lu2022fantastically}. 


\paragraph{Embedding-Based Adaptation.} Compared with weight-based approaches, embedding-based methods typically require far fewer parameters and do not modify model weights. For example, several representative embedding-based methods encode task information using roughly 0.1K--130K parameters for Llama-3.1-8B~\citep{todd2024function, peng2024live, huang2024multimodal, li2025implicit, park2025softinjectiontaskembeddings}, whereas LoRA adapters for the same model require more than 3M parameters. However, both ICL-derived and learned embedding methods require access to internal model activations to extract or inject task embeddings. ICL-derived approaches typically require precomputing task embeddings and identifying appropriate injection locations for each task~\citep{todd2024function, park2025softinjectiontaskembeddings}. In contrast, learned embedding approaches, including soft prompting methods, can suffer from optimization instability and sensitivity to embedding initialization and hyperparameter settings~\citep{lester-etal-2021-power, Li2021PrefixTuningOC, chen2022revisiting}. In addition, recent learned embedding methods often require thousands to tens of thousands of training iterations to converge~\citep{peng2024live, saglam2025learning, kang2025adaptivetaskvectorslarge, li2025towards}. 

Overall, Table~\ref{tab:adaptation_comparison} highlights the key characteristics that distinguish these paradigms and provides insight into why different task adaptation strategies have emerged. In Sections \ref{sec:weight-based adaptation}–\ref{sec:embedding-based adaptation}, we examine representative methods within each paradigm and discuss how these methods have evolved in light of the perspective summarized in Table~\ref{tab:adaptation_comparison}. As a supplementary reference, Appendix~\ref{sec:quantitative meta-analysis} provides an illustrative quantitative snapshot in Table~\ref{tab:bbh_grouped_results_params_runtime}, based on values reported in prior work.

\vspace{-1.5mm}
\section{Weight-Based Adaptation}
\label{sec:weight-based adaptation}
\vspace{-1.5mm}
Weight-based adaptation encodes task-relevant information in model weights or additional trainable modules. In this section, we categorize weight-based methods into single-task and multi-task adaptations and describe their respective methodologies in detail. Single-task adaptation primarily focuses on improving training efficiency and task performance, while multi-task adaptation addresses challenges in switching tasks during inference or performing batched inference where prompts in the same batch correspond to different tasks.

\subsection{Single-Task Adaptation}
\label{sec:single-task adaptation}
\paragraph{LoRA Variants. }LoRA~\citep{hu2022lora} introduces trainable low-rank updates to pretrained weight matrices. Due to its strong and stable performance, negligible inference overhead, and simple modular design, LoRA has inspired a large body of follow-up work that develops variants to improve expressivity, training dynamics, and efficiency. One line of work enhances the expressivity of model updates by increasing the intrinsic rank through alternative parametrizations, such as replacing standard matrix multiplication with Hadamard or Kronecker products~\citep{HyeonWoo2021FedParaLH, edalati2022kronaparameterefficienttuning, yeh2023navigating, huang2025hira}. Another line focuses on improving training stability and convergence speed through better initialization strategies, such as leveraging singular value decomposition of pretrained weights or gradient matrices to initialize LoRA modules more effectively~\citep{meng2024pissa, wang2024lora}. A further line of work explores adaptive rank allocation, dynamically adjusting the effective rank across LoRA modules during training~\citep{zhang2023adaptive}, or assigning different ranks to different modules based on their importance~\citep{ding-etal-2023-sparse, liu-etal-2024-alora}. Other variants target training-time memory efficiency, particularly for large-scale LLMs, by combining LoRA with low-bit quantization of pretrained weights during training~\citep{dettmers2023qlora, xu2024qalora, guo2024lqlora}. Complementary approaches further reduce the number of trainable parameters or storage requirements through parameter sharing across LoRA modules~\citep{kopiczko2024vera, renduchintala-etal-2024-tied, li2024vblora}, highlighting the flexibility of the LoRA framework. 

\paragraph{Other PEFT. }Beyond LoRA, a variety of PEFT methods have been proposed. Adapters~\citep{houlsby2019parameter} insert small trainable modules between Transformer layers of a pretrained model, while Compacter~\citep{mahabadi2021compacter} reduces adapter parameters by replacing adapter linear layers with low-rank hypercomplex multiplication. Other approaches reduce trainable parameters by selectively updating subsets of model weights. For example, FISH Mask~\citep{sung2021training} updates parameters based on their approximate Fisher Information~\citep{fisher1922mathematical, amari1996neural, kirkpatrick2017overcoming}, and BitFit~\citep{ben-zaken-etal-2022-bitfit} fine-tunes only bias terms. Another line of work applies element-wise rescaling operations to internal activations. (IA)$^3$~\citep{liu2022fewshot} introduces learnable scaling vectors that rescales internal activations, RED~\citep{wu-etal-2024-advancing} extends this idea by jointly learning scaling and bias vectors, and PARA~\citep{liu2024parameter} trains a lightweight vector generator to dynamically produce scaling vectors. Some methods combine or extend PEFT techniques, such as DePT~\citep{shi2024dept}, which integrates prompt tuning with LoRA-style updates on frozen word embeddings, and its variant ADePT~\citep{tang2025adept}. Other approaches focus on improving efficiency. LST~\citep{sung2022lst} trains a small auxiliary network to reduce training-time memory requirements by avoiding backpropagation through the backbone model, and RoAD~\citep{liao20243} applies trainable rotations to linear layer outputs to reduce the trainable parameters.

\subsection{Multi-Task Adaptation}
\label{sec:multi-task adaptation}
\paragraph{Multi-Task PEFT. }Compared to prompt-based and embedding-based adaptation, weight-based methods typically incur higher overhead when switching tasks during inference or processing mixed-task batches, as task-specific modules must be dynamically loaded and scheduled across requests. To mitigate this limitation, several methods introduce routing-based mechanisms that maintain multiple adapters and use (soft)~routers to select or combine task-relevant adapters, either at the task level~\citep{pfeiffer2021adapterfusion, huang2023lorahub, wang2023customizable} or at the instance level~\citep{wang2023multilora, feng2024mixture, tian2024hydralora, xu2024meteora, liu2025r}. These routers are typically implemented using learnable scalar weights or shallow neural networks. Other approaches explore alternative strategies for multi-task inference. O-LoRA~\citep{wang2023orthogonal} proposes a LoRA-based continual learning framework, while FLoRA~\citep{wen2024batched} replaces LoRA updates with Hadamard-product-based updates, enabling efficient heterogeneous batching by avoiding expensive batched matrix multiplications.

\paragraph{Model Merging. }Model merging defines \emph{task vectors}~\citep{ilharco2023editing} as parameter differences between fine-tuned models and a shared pretrained model and combines multiple such vectors to enable multi-task inference without additional fine-tuning. Although model merging typically underperforms multi-task training, it is particularly useful when fine-tuned models are available but the corresponding training data cannot be accessed, for example due to data privacy or intellectual property constraints. Several methods aim to reduce interference between task-specific parameter updates during model merging~\citep{yadav2023tiesmerging, yu2024language, gargiulo2025task}, while others propose fine-grained merging strategies that assign different merging coefficients across tasks, layers, or parameters~\citep{yang2024adamerging, zhou-etal-2024-metagpt, xu2025scalable, yao2025activationguided, lee-etal-2025-dynamic}. Another line of work focuses on performing model merging only on PEFT updates~\citep{zhang2023composing, tang2023parameter, zhao2024merging}. Among these, LoRA-LEGO~\citep{zhao2024merging} introduces rank-wise LoRA merging by identifying row-column pairs of LoRA matrices as minimal semantic units, clustering them across tasks, and using cluster centroids to construct merged adapters.
\section{Prompt-Based Adaptation}
\label{sec:prompt-based adaptation}
Prompt-based adaptation, commonly referred to as in-context learning~(ICL), adapts LLMs by providing task-relevant information directly in the input prompt at inference time, in the form of natural-language instructions and/or input-output demonstrations. Existing methods can be broadly categorized into \emph{instruction optimization}, which focuses on refining task instructions, and \emph{demonstration optimization}, which selects, orders, or generates input-output demonstrations. Collectively, these methods primarily aim to improve stability and performance of prompt-based adaptation.

\subsection{Instruction Optimization}
\label{sec:instruction optimization}
Instruction optimization focuses on refining task-specific instructions used in the prompt. These methods typically adopt a two-LLM framework, consisting of an \emph{optimizer LLM} that updates instructions and a \emph{task LLM} that performs the downstream task using the updated instructions. In practice, strong proprietary models such as GPT-4o~\citep{hurst2024gpt} are often used as the optimizer LLM to enable more effective instruction optimization. A common strategy is to iteratively refine instructions using feedback generated by the optimizer LLM itself~\citep{Pryzant2023AutomaticPO, wang2024promptagent, agarwal-etal-2025-promptwizard, juneja2025task, yan2025efficient}. For example, ProTeGi~\citep{Pryzant2023AutomaticPO} uses the optimizer LLM to analyze errors made under the current instruction and feeds the resulting feedback back into the optimizer LLM to update the instructions. Other approaches formulate instruction optimization as an evolutionary process, iteratively improving instructions through mutation, selection, or related evolutionary operations~\citep{guo2024connecting, fernando2024promptbreeder, agrawal2026gepa}. For example, GEPA~\citep{agrawal2026gepa} combines reflective feedback with evolutionary search, using natural-language reflection to propose prompt updates and Pareto-based selection to retain promising candidates.
Several methods leverage carefully designed meta-prompts, i.e., prompts that instruct the optimizer LLM how to refine instructions, to improve optimization stability and effectiveness~\citep{ye2024prompt, tang2025unleashing}. 
In contrast, some approaches avoid directly optimizing instructions and instead optimize auxiliary components, such as soft prompts~\citep{lester-etal-2021-power} provided to the optimizer LLM or the optimizer LLM itself~\citep{pmlr-v235-chen24e, lin2024instinct, kwon-etal-2024-stableprompt, zhao2025how}. For instance, InstructZero~\citep{pmlr-v235-chen24e} and INSTINCT~\citep{lin2024instinct} optimize only soft prompts, while StablePrompt~\citep{kwon-etal-2024-stableprompt} fine-tunes the optimizer LLM to improve instruction generation.

Beyond the optimization of task instructions, recent work on LLM agents has explored the automatic construction and refinement of \emph{agent skills}, which are reusable artifacts that encode procedural knowledge and tool-use strategies for a particular task family or domain~\citep{mi2026skill, alzubi2026evoskill, ni2026trace2skill, yang2026skillopt}. From the perspective of prompt-based adaptation, methods that refine the textual content of agent skills are closely related to instruction optimization: both adapt the task LLM without updating its parameters by modifying natural-language content provided in the input prompt, often using feedback from black-box task evaluations. The distinction between them therefore lies primarily in the type of content being optimized: instruction optimization refines instructions specifying a task's objective, constraints, expected output, or solution strategy, whereas agent-skill methods construct or refine reusable procedural knowledge that guides how an agent acts over multiple steps, interacts with tools, and responds to intermediate outcomes.

\subsection{Demonstration Optimization}
\label{sec:demonstration optimization}
Including a small number of input-output demonstrations in the prompt can effectively convey task information to LLMs~\citep{brown2020language}. However, task performance is highly sensitive to both which demonstrations are selected and how they are ordered within the prompt~\citep{zhao2021calibrate, liu2022makes, lu2022fantastically}. As a result, a large body of work has explored methods for optimizing demonstration selection and ordering. KATE~\citep{liu2022makes} retrieves demonstrations by embedding candidate examples and selecting nearest neighbors for each test input, and subsequent methods such as MDL~\citep{wu2023self} and ConE~\citep{peng-etal-2024-revisiting} adopt this retrieval step to first narrow the candidate pool before applying more refined selection strategies. Another line of work explicitly promotes diversity among selected demonstrations using determinantal point processes~(DPPs)~\citep{kulesza2012determinantal, ye2023compositional, yang2023representative}. Beyond selecting which demonstrations to include, several methods also consider demonstration ordering. Some approaches treat an ordered sequence of demonstrations as the basic unit of selection rather than scoring demonstrations independently~\citep{lu2022fantastically, wu2023self, wu2024prompt, guo2024makes}, while others formulate demonstration selection as a sequential decision-making process optimized with reinforcement learning~\citep{zhang-etal-2022-active, scarlatos2023reticl, wang2025demonstration}. For example, AES~\citep{zhang-etal-2022-active} formulates demonstration selection as a Markov Decision Process, where the policy state consists of all previously selected demonstrations, and trains an offline Q-learning policy~\citep{mnih2013playing} to select each new demonstration conditioned on this history.

In settings where only unlabeled data are available, several methods selectively annotate demonstrations to support in-context learning~\citep{hongjin2022selective, mavromatis2023examplesannotateincontextlearning, zhang2024ideal}. For example, Vote-K~\citep{hongjin2022selective} annotates diverse unlabeled examples using a $k$-nearest neighbor graph constructed in the Sentence-BERT~\citep{reimers2019sentence} embedding space, while IDEAL~\citep{zhang2024ideal} extends this approach with influence-driven selection to better approximate the underlying data distribution. More recently, LLM-generated demonstrations have emerged as an effective alternative to human-curated examples~\citep{long2024llms, yehudai2024genie, nadas2025synthetic}. Some approaches generate synthetic input-output pairs directly~\citep{chen-etal-2023-self, su2024demonstration, wan2025from}, while others augment demonstrations with LLM-generated reasoning paths~\citep{shao2023synthetic, he2024using, honda-oka-2025-exploring}, enabling chain-of-thought~(CoT)~\citep{wei2022chain} reasoning during inference. When a large number of demonstrations are available, scaling the number of demonstrations from a few to many, a setting commonly referred to as \emph{many-shot ICL}, can further improve task performance and robustness to demonstration selection and ordering~\citep{zhang-etal-2025-selecting, bertsch2025context}. However, these gains come at the cost of increased memory and computation that scale with the number of demonstrations, as well as attention dispersion~\citep{yuan-etal-2024-focused} in long prompts, which can degrade performance. 
\section{Embedding-Based Adaptation}
\label{sec:embedding-based adaptation}
Embedding-based adaptation represents task information as explicit vectors, referred to as \emph{task embeddings}, which are injected into model activations during inference, typically via addition or concatenation. Existing methods can be broadly divided into two classes based on how these embeddings are obtained: \emph{ICL-derived task embeddings} and \emph{learned task embeddings}. The former extracts task embeddings from internal activations induced by ICL, while the latter directly optimizes task embeddings through gradient-based training. Although these two classes are respectively related to ICL and PEFT, both aim to improve downstream task performance without relying on task-specific prompts during inference.

\subsection{ICL-Derived Task Embeddings}
\label{sec:icl-derived task embeddings}

\paragraph{Non-Contrastively Derived. }Early work, notably Task Vectors~\citep{ hendel-etal-2023-context} and Function Vectors~\citep{todd2024function}, showed that last-token internal activations produced during few-shot inference encode rich task-relevant information, which can be extracted and injected into model activations to enable task execution in a zero-shot setting. Building on this observation, non-contrastive embedding-based adaptation methods aggregate such activations across multiple few-shot prompts and aim to identify effective injection locations that maximize task performance. Injection locations are determined using various strategies, including validation-based sweeps over layers~\citep{hendel-etal-2023-context, zhang-etal-2024-batch, wang2025elicit}, reinforcement-learning-based optimization~\citep{huang2024multimodal, cai-etal-2025-beyond}, and gradient-based optimization that softly controls task injection~\citep{li2025implicit, park2025softinjectiontaskembeddings}. While most approaches operate at the layer level, some explore finer-grained injection at the attention-head level~\citep{huang2024multimodal, park2025softinjectiontaskembeddings}. Recent results indicate that these methods can achieve performance comparable to few-shot ICL; however, evaluations have largely focused on relatively simple tasks, motivating broader evaluation on more complex reasoning and generation tasks.

\paragraph{Contrastively Derived. }Contrastively derived task embeddings originate from Inference-Time Intervention~(ITI)~\citep{li2023inferencetime}, which showed that differences between last-token activations induced by contrastive prompt pairs encode directional information that can be injected into model activations to steer model behavior. ITI and subsequent work have leveraged such contrastive prompt pairs primarily for behavior steering, including improving truthfulness or safety ~\citep{liu2024context, rimsky-etal-2024-steering, qiu2024spectral, wang2025adaptive, zhan2025realreadingtransformeractivations}. More recent methods extend this paradigm to task adaptation by constructing contrastive pairs where one prompt contains correct task demonstrations and the other contains incorrect or missing demonstrations, enabling task execution via the resulting difference vectors~\citep{ wang2025semanticsadaptive, liu2025iterative}. Using contrastively derived embeddings for general task adaptation is relatively underexplored, and systematic comparisons with non-contrastive approaches remain an open area for further study.

\subsection{Learned Task Embeddings}
\label{sec:learned task embeddings}

\paragraph{Soft Prompting. }Prompt Tuning~\citep{lester-etal-2021-power} adapts models by prepending learnable continuous embeddings, known as \emph{soft prompts}, to the input embedding sequence and optimizing only these embeddings. Although highly parameter-efficient, prompt tuning is often unstable and typically underperforms weight-based PEFT methods such as LoRA~\citep{hu2022lora, liu2022fewshot}, motivating subsequent improvements. Some approaches dynamically adapt task-specific soft prompts to produce instance-specific soft prompts, leading to improved performance~\citep{asai-etal-2022-attempt, jain2024prompt}. Another line of work explores transfer learning by leveraging soft prompts from multiple source tasks to initialize or compose target-task prompts, thereby transferring task knowledge and improving adaptation~\citep{vu-etal-2022-spot, asai-etal-2022-attempt, wang2023multitask, belanec2025task, zhang2025dynamic}. For example, SPoT~\citep{vu-etal-2022-spot}, MPT~\citep{wang2023multitask}, and Task Prompt Vectors~\citep{belanec2025task} initialize target-task soft prompts using soft prompts learned from related source tasks.

\paragraph{Other Learned Embeddings. }Other learned-embedding approaches adapt models by optimizing task embeddings via gradient-based training, where the embeddings are injected into internal activations, typically via additive intervention. LIVE~\citep{peng2024live} and M$^2$IV~\citep{li2025miv} introduce layer-wise learnable vectors with associated scaling factors that are trained end-to-end. LTV~\citep{saglam2025learning} extracts attention-head outputs from few-shot inference and constructs layer-wise task embeddings as learned weighted combinations of heads within each layer. However, these methods often require a large number of training iterations, sometimes exceeding those of standard PEFT approaches~\citep{peng2024live, saglam2025learning, kang2025adaptivetaskvectorslarge, li2025towards}. Moreover, these methods are often compared with ICL-derived embedding approaches, while their conceptual similarity to soft prompting methods is often overlooked.
\section{Open Problems and Discussion}
\label{sec: Open Problems}
In this section, we present four open research problems that cut across all three task adaptation paradigms for LLMs. Additional discussions are provided in Appendix~\ref{sec:discussion}, including comparisons between our taxonomy and prior PEFT-only or ICL-only survey taxonomies, the distinction between weight-based and embedding-based adaptations, the rationale for classifying soft prompting as embedding-based adaptation, and a limited quantitative analysis.

\paragraph{Hybrid and Compositional Task Adaptation. }Beyond relying on a single adaptation paradigm, an important research problem is to jointly leverage multiple forms of task information, such as learned embeddings, prompts, and trained modules, to strengthen task adaptation. As a representative example, Instruction Prompt Tuning~\citep{singhal2023large} prepends a shared soft prompt to task-specific instructions and demonstrations, enabling complementary task signals to jointly guide model behavior and better align an instruction-tuned LLM with domain-specific instruction semantics, thereby improving the safety, grounding, and completeness of long-form generations. Despite such initial efforts, how best to combine different task adaptation paradigms remains underexplored, presenting an important direction for future research.

\paragraph{Data Efficiency and Synthetic Supervision. }Most task adaptation methods require a nontrivial number of labeled demonstrations to achieve strong performance~\citep{lester-etal-2021-power, hu2022lora, agarwal2024manyshot}. To reduce reliance on human-curated supervision, LLM-generated data have emerged as a scalable alternative and are increasingly adopted for task adaptation~\citep{long2024llms, yehudai2024genie, nadas2025synthetic}. Prior work shows that synthetic demonstrations can differ from human-curated ones in fidelity, bias, and diversity~\citep{li2024data, nadas2025synthetic}, yet how these differences affect different task adaptation paradigms has not been thoroughly studied. Addressing this question could provide useful insights into how synthetic data should be used across different task adaptation paradigms.

\paragraph{Benchmarks for Evaluating Task Adaptation.}
Existing task adaptation methods have been evaluated on a broad range of established benchmarks, summarized in Tables~\ref{tab:benchmark (part 1 of 4)}-\ref{tab:benchmark (part 4 of 4)}. These benchmarks cover diverse categories, including natural language understanding, reasoning, mathematics, coding, summarization, and safety and trustworthiness. While such benchmarks have served as controlled testbeds for comparing adaptation methods, recent developments in LLM research suggest the need to expand evaluations to more challenging settings. These include generating long-form responses~\citep{wu2024longgenbench, bai2024longwriter, que2024hellobench}, solving complex multi-step reasoning problems~\citep{he2024olympiadbench, glazer2024frontiermath, phan2025humanity, aops_2025_aime_i, lin2025zebralogic, balunovic2025matharena}, and maintaining coherent behavior in long-horizon or interactive environments~\citep{shridhar2020alfworld, wang-etal-2022-scienceworld, yao2022webshop, zhou2023webarena, wei2025browsecomp}. Evaluating adaptation methods under such settings may provide deeper insights into their robustness, error accumulation, and long-range reasoning capabilities.

\paragraph{Beyond Task Adaptation: Encoding Behavioral and Contextual Information. } Beyond task adaptation, these techniques have also been explored for encoding diverse behavioral and contextual signals in LLMs, including safety/truthfulness, persona, and long-context knowledge. Prompting offers a simple and flexible interface for specifying such signals in natural language, but its effectiveness can be limited for some attributes~(e.g., safety or persona)~\citep{zheng2024prompt, zheng2024helpful}, and long prompts may introduce additional inference overhead. To address these limitations, recent methods encode such information directly into model parameters or embeddings. For example, SHINE~\citep{liu2026shine} and Doc-to-LoRA~\citep{charakorn2026doc} generate LoRA adapters that internalize information from long-contexts. More broadly, embedding-based approaches have been used to encode various types of information, including safety/truthfulness~\citep{li2023inferencetime, wang2025adaptive}, persona~\citep{poterti2025can, feng2026persona}, and long-context knowledge~\citep{eyuboglu2025cartridges}, suggesting that expanding the range of information encoded through such adaptation techniques is an interesting research direction.

\section{Conclusion}
This survey presents a unified overview of task adaptation in LLMs by organizing existing methods according to where task information is encoded. Through this framework, we connect weight-based, prompt-based, and embedding-based adaptations and discuss their relationships, trade-offs, and shared open problems. We hope this survey serves as a useful reference for understanding the independently evolving landscape of LLM task adaptation.
\section*{Limitations}
\label{sec:Limitations}
This survey provides a unified overview of task adaptation techniques for large language models, but it has several limitations. First, given the rapid pace of recent progress in this area, the taxonomy and comparisons presented here may not fully capture the most recent developments. Second, the survey emphasizes high-level methodological distinctions and qualitative trade-offs rather than exhaustive empirical re-evaluation of individual methods. This focus is intended to highlight the key strengths and limitations of different adaptation paradigms, thereby contextualizing existing approaches and motivating future research. Finally, although many task adaptation techniques for LLMs may be applicable to other types of models, such as multimodal LLMs or visual generative models, this survey focuses exclusively on text-based LLMs, leaving such extensions for future work.
\section*{Acknowledgments}

This work was supported by Institute of Information \& communications Technology Planning \& Evaluation (IITP) grant funded by the Korea government (MSIT) ([NO.RS-2021-II211343, Artificial Intelligence Graduate School Program (Seoul National University)], [No.RS-2023-00235293, Development of autonomous driving big data processing, management, search, and sharing interface technology to provide autonomous driving data according to the purpose of usage]) and the InnoCORE program of the Ministry of Science and ICT (26-InnoCORE-01).

\bibliography{bibliography}

\clearpage
\appendix

\addtocontents{toc}{\protect\setcounter{tocdepth}{2}}

\section{Full Taxonomy of Efficient Task Adaptation in LLMs}
\label{sec:taxonomy_appendix}

\definecolor{rootgreen}{RGB}{208,248,206}
\definecolor{methodtan}{RGB}{255,242,204}
\definecolor{leafwhite}{RGB}{255,255,255}
\definecolor{softteal}{RGB}{200,235,235}   
\definecolor{iceblue}{RGB}{225,240,248}    
\definecolor{softgold}{RGB}{245,235,200}   
\definecolor{lightmustard}{RGB}{242,230,185} 
\definecolor{sagegreen}{RGB}{215,230,205}  
\definecolor{mintgreen}{RGB}{205,240,225}  
\definecolor{warmgray}{RGB}{235,235,235}   
\definecolor{coolgray}{RGB}{225,230,235}   
\definecolor{softviolet}{RGB}{225,210,245} 
\definecolor{dustypurple}{RGB}{215,205,230} 
\definecolor{lavender}{RGB}{235,210,235}   
\definecolor{gray}{RGB}{220,220,220}   
\definecolor{palered}{RGB}{255,220,220}  
\definecolor{softcoral}{RGB}{255,210,200}  
\definecolor{softorange}{RGB}{255,225,200} 

\definecolor{coolgray}{RGB}{225,230,235}

\definecolor{softteal}{RGB}{200,235,235}
\definecolor{softtealA}{RGB}{220,245,245}  
\definecolor{softtealB}{RGB}{235,250,250}  

\definecolor{catblue}{RGB}{218,232,252}
\definecolor{catblueA}{RGB}{232,242,252}  
\definecolor{catblueB}{RGB}{242,248,254}  

\definecolor{lilacgray}{RGB}{232,228,240}  
\definecolor{lilacgrayA}{RGB}{242,240,248}
\definecolor{lilacgrayB}{RGB}{248,247,252}

\definecolor{softviolet}{RGB}{225,210,245}
\definecolor{softvioletA}{RGB}{235,225,250}
\definecolor{softvioletB}{RGB}{245,238,252}

\definecolor{softcoral}{RGB}{255,210,200}
\definecolor{softcoralA}{RGB}{255,225,218}
\definecolor{softcoralB}{RGB}{255,238,232}

\definecolor{sagegreen}{RGB}{215,230,205}
\definecolor{softgold}{RGB}{245,235,200}

\definecolor{embcol}{RGB}{210,230,255}   
\definecolor{parcol}{RGB}{255,230,190}   
\definecolor{procol}{RGB}{215,245,220}   

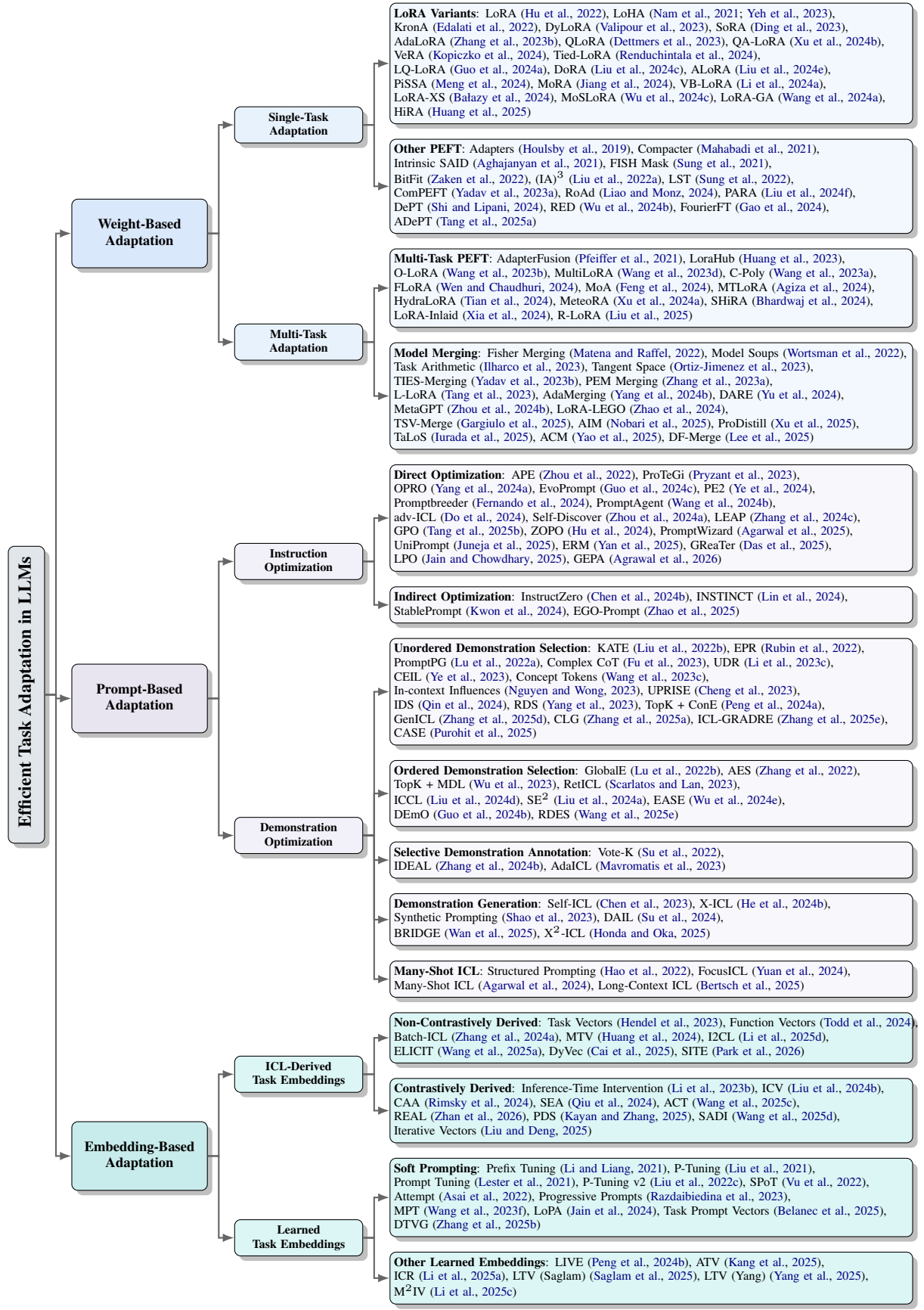
\begin{figure*}
\centering
\begin{forest}
  tikz={scale=0.68, transform shape},
  for tree={
    grow=east,
    forked edges,
    draw,
    rounded corners=3pt,
    align=center,
    edge={draw=black!60, -{Latex[length=2mm]}, line width=0.8pt},
    parent anchor=east,
    child anchor=west,
    anchor=west,
    drop shadow,
  },
  root style/.style={
    fill=coolgray,
    fit=band,
    font=\bfseries\small,
    inner sep=5pt,
    align=center,
  },
  levelone/.style={
    fill=catblue,
    fit=band,
    font=\bfseries\scriptsize,
    text width=2.3cm,
    inner sep=1pt,
    node options={align=center},
    minimum height=1.2cm,
  },
  leveltwo/.style={
    fill=methodtan,
    font=\tiny,
    text width=2.1cm,
    node options={align=center},
    inner sep=2pt,
  },
  levelthree/.style={
    fill=methodtan,
    font=\fontsize{6.2}{7.0}\selectfont,
    text width=\dimexpr\textwidth-7.0cm\relax,
    align=left,
    inner sep=2pt,
  },
  longleveltwo/.style={
    fill=methodtan,
    font=\fontsize{6.2}{7.0}\selectfont,
    text width=\dimexpr\textwidth-4.37cm\relax,
    align=left,
    inner sep=2pt,
  },
  leaf/.style={
    fill=leafwhite,
    font=\scriptsize,
    align=left,
    inner sep=4pt,
  },
  emb branch/.style={
    fill=softteal,
    draw
  },
  emb subbranchA/.style={
    fill=softtealA,
    draw
  },
  emb subbranchB/.style={
    fill=softtealA,
    draw
  },
  par branch/.style={
    fill=catblue,
    draw
  },
  par subbranchA/.style={
    fill=catblueA,
    draw
  },
  par subbranchB/.style={
    fill=catblueB,
    draw
  },
  pro branch/.style={
    fill=lilacgray,
    draw
  },
  pro subbranchA/.style={
    fill=lilacgrayA,
    draw
  },
  pro subbranchB/.style={
    fill=lilacgrayB,
  },
  [{\rotatebox{90}{Efficient Task Adaptation in LLMs}}, root style
    [Embedding-Based\\\textbf{Adaptation}, levelone, emb branch
        [\textbf{Learned}\\\textbf{Task Embeddings}, leveltwo, emb subbranchA
            [{\textbf{Other Learned Embeddings}: LIVE~\citep{peng2024live}, ATV~\citep{kang2025adaptivetaskvectorslarge}, \\
            ICR~\citep{li2025towards}, LTV (Saglam)~\citep{saglam2025learning}, LTV (Yang)~\citep{yang2025taskvectorslearnedextracted}, \\
            M$^2$IV~\citep{li2025miv}
            }, levelthree, emb subbranchB]
            [{\textbf{Soft Prompting}: Prefix Tuning~\citep{Li2021PrefixTuningOC}, P-Tuning~\citep{liu2021gpt}, \\
            Prompt Tuning~\citep{lester-etal-2021-power}, P-Tuning v2~\citep{liu-etal-2022-p}, SPoT~\citep{vu-etal-2022-spot}, \\
            Attempt~\citep{asai-etal-2022-attempt}, Progressive Prompts~\citep{razdaibiedina2023progressive},  \\
            MPT~\citep{wang2023multitask}, LoPA~\citep{jain2024prompt}, Task Prompt Vectors~\citep{belanec2025task}, \\
            DTVG~\citep{zhang2025dynamic} 
            }, levelthree, emb subbranchB]
        ] 
        [\textbf{ICL-Derived}\\\textbf{Task Embeddings}, leveltwo, emb subbranchA
            [{\textbf{Contrastively Derived}: Inference-Time Intervention~\citep{li2023inferencetime}, ICV~\citep{liu2024context},\\
            CAA~\citep{rimsky-etal-2024-steering}, SEA~\citep{qiu2024spectral}, ACT~\citep{wang2025adaptive}, \\ REAL~\citep{zhan2025realreadingtransformeractivations}, PDS~\citep{kayan2025prototypebaseddynamicsteeringlarge}, SADI~\citep{wang2025semanticsadaptive}, \\
            Iterative Vectors~\citep{liu2025iterative}
            }, levelthree, emb subbranchB]
            [{\textbf{Non-Contrastively Derived}: Task Vectors~\citep{hendel-etal-2023-context}, Function Vectors~\citep{todd2024function}, \\
            Batch-ICL~\citep{zhang-etal-2024-batch}, MTV~\citep{huang2024multimodal}, I2CL~\citep{li2025implicit}, \\
            ELICIT~\citep{wang2025elicit}, DyVec~\citep{cai-etal-2025-beyond}, SITE~\citep{park2025softinjectiontaskembeddings}
            }, levelthree, emb subbranchB]
        ]
    ]
    [Prompt-Based\\Adaptation, levelone, pro branch
       [\textbf{Demonstration}\\\textbf{Optimization}, leveltwo, pro subbranchA
          [{\textbf{Many-Shot ICL}: Structured Prompting~\citep{hao2022structuredpromptingscalingincontext}, FocusICL~\citep{yuan-etal-2024-focused},\\Many-Shot ICL~\citep{agarwal2024manyshot}, Long-Context ICL~\citep{bertsch2025context} 
          }, levelthree, pro subbranchB]
          [{\textbf{Demonstration Generation}: Self-ICL~\citep{chen-etal-2023-self}, X-ICL~\citep{he2024using}, \\
          Synthetic Prompting~\citep{shao2023synthetic}, DAIL~\citep{su2024demonstration}, 
          \\BRIDGE~\citep{wan2025from}, X$^2$-ICL~\citep{honda-oka-2025-exploring}
          }, levelthree, pro subbranchB]
          [{\textbf{Selective Demonstration Annotation}: Vote-K~\citep{hongjin2022selective}, \\IDEAL~\citep{zhang2024ideal}, 
          AdaICL~\citep{mavromatis2023examplesannotateincontextlearning}
          }, levelthree, pro subbranchB]
          [{\textbf{Ordered Demonstration Selection}: GlobalE~\citep{lu2022fantastically}, AES~\citep{zhang-etal-2022-active}, 
          \\TopK + MDL~\citep{wu2023self}, 
          RetICL~\citep{scarlatos2023reticl}, \\ICCL~\citep{liu2024let}, 
          SE$^2$~\citep{liu-etal-2024-se2}, EASE~\citep{wu2024prompt}, 
          \\DEmO~\citep{guo2024makes}, RDES~\citep{wang2025demonstration} 
          }, levelthree, pro subbranchB]
          [{\textbf{Unordered Demonstration Selection}: KATE~\citep{liu2022makes}, EPR~\citep{rubin2022learning}, 
          \\PromptPG~\citep{lu2022dynamic}, Complex CoT~\citep{fu2023complexitybased}, UDR~\citep{li-etal-2023-unified}, 
          \\CEIL~\citep{ye2023compositional}, Concept Tokens~\citep{wang2023large}, 
          \\In-context Influences~\citep{nguyen2023context}, UPRISE~\citep{cheng2023uprise}, 
          \\IDS~\citep{qin2024context}, RDS~\citep{yang2023representative}, TopK + ConE~\citep{peng-etal-2024-revisiting},  \\
          GenICL~\citep{zhang-etal-2025-learning-select}, CLG~\citep{zhang-etal-2025-selecting}, ICL-GRADRE~\citep{zhang2025lineartimedemonstrationselectionincontext}, \\
          CASE~\citep{purohit2025sample}
          }, levelthree, pro subbranchB]
      ]
      [\textbf{Instruction}\\\textbf{Optimization}, leveltwo, pro subbranchA
        [{\textbf{Indirect Optimization}: InstructZero~\citep{pmlr-v235-chen24e}, INSTINCT~\citep{lin2024instinct}, \\
        StablePrompt~\citep{kwon-etal-2024-stableprompt}, EGO-Prompt~\citep{zhao2025how}
        }, levelthree, pro subbranchB]
        [{\textbf{Direct Optimization}: APE~\citep{zhou2022large}, ProTeGi~\citep{Pryzant2023AutomaticPO}, 
        \\OPRO~\citep{yang2023large}, EvoPrompt~\citep{guo2024connecting}, PE2~\citep{ye2024prompt}, 
        \\Promptbreeder~\citep{fernando2024promptbreeder}, PromptAgent~\citep{wang2024promptagent}, 
        \\adv-ICL~\citep{do2024prompt}, Self-Discover~\citep{zhou2024selfdiscover}, LEAP~\citep{zhang2024incontext}, 
        \\GPO~\citep{tang2025unleashing}, ZOPO~\citep{hu2024localized}, PromptWizard~\citep{agarwal-etal-2025-promptwizard}, 
        \\UniPrompt~\citep{juneja2025task}, ERM~\citep{yan2025efficient}, GReaTer~\citep{das2025greater}, 
        \\LPO~\citep{jain-chowdhary-2025-local}, GEPA~\citep{agrawal2026gepa}
        }, levelthree, pro subbranchB]
      ]
    ]
    [Weight-Based\\Adaptation, levelone, par branch
        [\textbf{Multi-Task}\\\textbf{Adaptation}, leveltwo, par subbranchA
            [{\textbf{Model Merging}: Fisher Merging~\citep{matena2022merging}, Model Soups~\citep{pmlr-v162-wortsman22a}, \\
            Task Arithmetic~\citep{ilharco2023editing}, Tangent Space~\citep{ortiz2023task},  \\
            TIES-Merging~\citep{yadav2023tiesmerging}, PEM Merging~\citep{zhang2023composing},  \\
            L-LoRA~\citep{tang2023parameter}, AdaMerging~\citep{yang2024adamerging}, DARE~\citep{yu2024language}, \\
            MetaGPT~\citep{zhou-etal-2024-metagpt},  LoRA-LEGO~\citep{zhao2024merging},   \\
            TSV-Merge~\citep{gargiulo2025task}, AIM~\citep{nobari2025activationinformedmerginglargelanguage}, ProDistill~\citep{xu2025scalable},   \\ TaLoS~\citep{iurada2025efficient}, ACM~\citep{yao2025activationguided}, DF-Merge~\citep{lee-etal-2025-dynamic}
            }, levelthree, par subbranchB]
            [{\textbf{Multi-Task PEFT}: AdapterFusion~\citep{pfeiffer2021adapterfusion}, LoraHub~\citep{huang2023lorahub}, 
            \\O-LoRA~\citep{wang2023orthogonal}, MultiLoRA~\citep{wang2023multilora}, C-Poly~\citep{wang2023customizable}, 
            \\FLoRA~\citep{wen2024batched}, MoA~\citep{feng2024mixture}, MTLoRA~\citep{agiza2024mtlora},   \\
            HydraLoRA~\citep{tian2024hydralora}, MeteoRA~\citep{xu2024meteora}, SHiRA~\citep{bhardwaj2024sparse},  \\
            LoRA-Inlaid~\citep{xia2024efficient}, R-LoRA~\citep{liu2025r}
            }, levelthree, par subbranchB]
        ]
        [\textbf{Single-Task}\\\textbf{Adaptation}, leveltwo, par subbranchA
            [{\textbf{Other PEFT}: Adapters~\citep{houlsby2019parameter}, Compacter~\citep{mahabadi2021compacter}, \\
            Intrinsic SAID~\citep{aghajanyan-etal-2021-intrinsic}, FISH Mask~\citep{sung2021training}, \\
            BitFit~\citep{ben-zaken-etal-2022-bitfit}, (IA)$^3$~\citep{liu2022fewshot}, LST~\citep{sung2022lst},  \\
            ComPEFT~\citep{yadav2023compeft}, RoAd~\citep{liao20243}, PARA~\citep{liu2024parameter}, \\
            DePT~\citep{shi2024dept}, RED~\citep{wu-etal-2024-advancing}, FourierFT~\citep{gao2024parameter}, \\ 
            ADePT~\citep{tang2025adept}
            }, levelthree, par subbranchB]
            [{\textbf{LoRA Variants}: LoRA~\citep{hu2022lora}, LoHA~\citep{HyeonWoo2021FedParaLH,yeh2023navigating}, \\
            KronA~\citep{edalati2022kronaparameterefficienttuning}, DyLoRA~\citep{valipour-etal-2023-dylora}, SoRA~\citep{ding-etal-2023-sparse}, \\
            AdaLoRA~\citep{zhang2023adaptive}, QLoRA~\citep{dettmers2023qlora}, QA-LoRA~\citep{xu2024qalora},\\
            VeRA~\citep{kopiczko2024vera}, Tied-LoRA~\citep{renduchintala-etal-2024-tied}, \\
            LQ-LoRA~\citep{guo2024lqlora}, DoRA~\citep{liu2024dora}, ALoRA~\citep{liu-etal-2024-alora},  \\
            PiSSA~\citep{meng2024pissa}, MoRA~\citep{jiang2024mora}, VB-LoRA~\citep{li2024vblora}, \\
            LoRA-XS~\citep{balazy2024lora}, MoSLoRA~\citep{wu-etal-2024-mixture-subspaces}, LoRA-GA~\citep{wang2024lora}, \\ 
            HiRA~\citep{huang2025hira}
            }, levelthree, par subbranchB]
        ]
    ]
  ]
\end{forest}
\caption{\textbf{Full Taxonomy of Efficient Task Adaptation in Large Language Models.}}
\label{fig:taxonomy}
\end{figure*}

Figure~\ref{fig:taxonomy} presents the full taxonomy of efficient task adaptation methods in large language models. In contrast to the reduced taxonomy in Figure~\ref{fig:taxonomy_reduced} of Section~\ref{sec:introduction}, which shows only the main category branches, the full taxonomy includes the individual methods associated with each subcategory.

\section{Discussion}
\label{sec:discussion}
\subsection{Comparison with Previous PEFT-only and ICL-only Taxonomies}
This work proposes a unified taxonomy that covers weight-based, prompt-based, and embedding-based task adaptation techniques. While most embedding-based adaptation methods, except for soft prompting~\citep{lester-etal-2021-power, Li2021PrefixTuningOC}, have only recently emerged and therefore lack dedicated surveys, several PEFT-only and ICL-only surveys have previously proposed taxonomies for weight-based and prompt-based adaptation methods~\citep{lialin2023scaling, dong2024survey, sahoo2024systematic, vatsal2024survey, han2024parameter, wang2025parameter, mao2025survey}. In this section, we compare our taxonomy with those proposed in representative PEFT-only and ICL-only surveys. We select representative surveys based on citation impact, relevance to each subfield, and publication or arXiv release year. The selected surveys are listed below:

\paragraph{PEFT-only surveys}
\begin{itemize}
\item P1. \emph{Parameter-Efficient Fine-Tuning for Large Models: A Comprehensive Survey}~\citep{han2024parameter}
\item P2. \emph{Parameter-Efficient Fine-Tuning in Large Models: A Survey of Methodologies}~\citep{wang2025parameter}
\item P3. \emph{A Survey on LoRA of Large Language Models}~\citep{mao2025survey}
\end{itemize}

\paragraph{ICL-only surveys}
\begin{itemize}
\item I1. \emph{A Survey on In-context Learning}~\citep{dong2024survey}
\item I2. \emph{A Systematic Survey of Prompt Engineering in Large Language Models: Techniques and Applications}~\citep{sahoo2024systematic}
\item I3. \emph{A Survey of Prompt Engineering Methods in Large Language Models for Different NLP Tasks}~\citep{vatsal2024survey}
\end{itemize}

For ease of discussion, we refer to these surveys as P1--P3 for PEFT surveys and I1--I3 for ICL surveys. It is important to note that taxonomies proposed in different surveys naturally reflect the scope and objectives of the respective works. Our goal here is therefore not to claim that our taxonomy is generally preferable to previous ones, but to clarify how the coverage and perspective adopted in our taxonomy differ from prior PEFT-only and ICL-only taxonomies and how they align with the goal of contextualizing a broad range of methods under the theme of task adaptation in LLMs.

Compared with previous single-field surveys, our taxonomy differs in two main aspects. First, our taxonomy spans weight-based, prompt-based, and embedding-based adaptation within a single framework. This broader coverage provides a unified perspective on task adaptation in LLMs and enables a contextualized view of different approaches based on how task information is encoded and utilized. Second, most existing PEFT-only and ICL-only surveys primarily cover works published up to around 2024, whereas our survey additionally includes more recent methods published in 2025 and 2026. Considering that several directions, such as instruction optimization, have recently evolved, introducing a new taxonomy helps better reflect the current landscape of task adaptation methods. Below, we further compare our taxonomy with previous ones for weight-based and prompt-based adaptation, respectively.

\paragraph{Detailed Comparison with PEFT-only Taxonomies. }
The PEFT-only surveys P1 and P2 categorize methods primarily according to fine-grained technical design choices, such as adapter-based methods, selective parameter updates, reparameterization-based fine-tuning, and quantization-based approaches. Similarly, P3 focuses on detailed categorization of LoRA variants. These taxonomies provide fine-grained methodological classifications that help organize the large body of PEFT research. In contrast, our taxonomy places PEFT methods within the broader category of \emph{weight-based adaptation}. This category additionally includes approaches such as model merging and multi-task adaptation, which are not typically emphasized in previous PEFT-only taxonomies. Instead of organizing methods by specific architectural techniques, our taxonomy categorizes subfields according to the challenges they address in task adaptation. Specifically, as discussed in Section~\ref{sec:weight-based adaptation}, weight-based adaptation is first divided into single-task and multi-task adaptation. Single-task adaptation primarily focuses on improving training efficiency and task performance, whereas multi-task adaptation addresses challenges such as switching tasks during inference or performing mixed-task batched inference. Within multi-task adaptation, we further distinguish between multi-task PEFT and model merging. While both approaches enable multi-task adaptation, model merging is particularly relevant in scenarios where training data are unavailable and only task-specific adapted models are accessible. Because these perspectives are not explicitly reflected in existing PEFT-only taxonomies, introducing a taxonomy tailored to the broader context of task adaptation is more suitable for a unified survey such as ours.

\paragraph{Detailed Comparison with ICL-only Taxonomies. }Previous ICL surveys adopt different perspectives in organizing prompt-based methods. The taxonomy in I1 covers both methodological developments and analytical studies of ICL. In contrast, I2 and I3 categorize prompt engineering techniques primarily according to their target applications, such as logical reasoning, hallucination reduction, emotion or tone control, and code generation. Our taxonomy organizes prompt-based adaptation from a different perspective by focusing on how task information is encoded and on the main challenges addressed in task adaptation. Under this perspective, our taxonomy includes categories such as unordered demonstration selection, ordered demonstration selection, selective demonstration annotation, demonstration generation, and many-shot ICL. Several of these directions are either not included or only partially covered in previous ICL surveys. For example, recent instruction optimization methods have become an important branch of prompt-based adaptation but are only sparsely represented in earlier surveys due to their recent emergence. Similarly, subfields such as selective demonstration annotation, demonstration generation, and many-shot ICL play important supporting roles in prompt-based task adaptation but are not systematically categorized in previous ICL taxonomies. By incorporating these directions, our taxonomy provides a more comprehensive contextualization of prompt-based task adaptation methods.

\paragraph{Summary. }Overall, our taxonomy differs from previous PEFT-only and ICL-only taxonomies in both scope and organizing perspective. While previous surveys focus on individual subfields and often categorize methods by architectural techniques or application domains, our taxonomy provides a unified view spanning weight-based, prompt-based, and embedding-based adaptation. By organizing methods according to the challenges they address in task adaptation, our taxonomy highlights conceptual connections across different adaptation strategies while preserving the methodological insights developed within each individual research area.

\subsection{Detailed Comparison of Weight-Based and Embedding-Based Adaptations}
In our taxonomy, we distinguish between weight-based and embedding-based adaptations. A conceptual comparison of these paradigms is summarized in Table~\ref{tab:adaptation_comparison} of Section~\ref{sec: A Unified View of Task Adaptation in Large Language Models}. In this section, we further clarify the differences between these two adaptation paradigms. 

Following terminology commonly used in prior literature~\citep{bishop2006pattern, goodfellow2016deep}, we first clarify the meanings of \emph{weights} and \emph{embeddings (or activations)}. Conceptually, weights are parameters that define the transformations applied to activations, typically involving multiplicative operations. In contrast, embeddings or activations refer to the outputs produced by these transformations. Based on this distinction, we categorize methods that encode task information into model weights, including additional weights or rescaling coefficients, as \emph{weight-based adaptation}. In contrast, methods that encode task information directly in activation-space representations are categorized as \emph{embedding-based adaptation}. This includes approaches that encode task information in soft prompts within the input embedding space or in injected activation vectors within the activation space of intermediate layers.

At first glance, this distinction may appear unfamiliar because weight-based PEFT methods such as LoRA~\citep{hu2022lora} are often discussed within the broader PEFT literature together with soft prompting methods. In our taxonomy, however, weight-based PEFT methods are categorized as weight-based adaptation because they encode task information in model weights, whereas soft prompting methods are categorized as embedding-based adaptation because task information is encoded in the embedding space. We introduce this categorization because it more clearly reflects where task information is encoded in the model. 

This perspective also helps clarify relationships between different adaptation approaches. For instance, recently proposed learned embedding methods~\citep{peng2024live, li2025miv, saglam2025learning, li2025towards, kang2025adaptivetaskvectorslarge, yang2025taskvectorslearnedextracted}, such as LTV~\citep{saglam2025learning}, share conceptual similarities with soft prompting approaches because both encode task information in activation-space representations optimized through gradient-based training. However, this connection is often overlooked in previous studies~\citep{peng2024live, saglam2025learning, li2025towards, yang2025taskvectorslearnedextracted}, which rarely discuss or compare these methods with soft prompting approaches. By organizing methods according to where and how task information is encoded, our taxonomy highlights these conceptual connections and provides a unified framework for systematically analyzing and comparing a broad range of task adaptation techniques.

\subsection{Quantitative Meta-Analysis}
\label{sec:quantitative meta-analysis}
The values summarized in Table~\ref{tab:bbh_grouped_results_params_runtime} are derived from the results reported in prior work~\citep{park2025softinjectiontaskembeddings}. We provide this focused quantitative comparison to complement the qualitative analysis in Section~\ref{sec: A Unified View of Task Adaptation in Large Language Models} by illustrating differences in task performance, trainable parameter count, and runtime among representative methods.

The Big-Bench Hard scores and trainable parameter counts illustrate the trade-off between task performance and parameter efficiency. Weight-based adaptation remains a strong baseline, with LoRA and (IA)$^3$ achieving the highest scores among the compared methods. In contrast, 10-shot ICL requires no trainable parameters but performs lower than these PEFT methods, reflecting the trade-off between training-free adaptation and task performance. Embedding-based methods use substantially fewer trainable parameters than weight-based methods. Among ICL-derived embedding methods, SITE achieves the strongest performance, outperforming 10-shot ICL while using only 1.02K trainable parameters. This suggests that ICL-derived task embeddings have the potential to provide a compact alternative to repeatedly including demonstrations in the input prompt. However, the low scores of FV and LIVE also indicate that embedding-based adaptation does not uniformly improve performance, and its effectiveness depends strongly on how task embeddings are constructed and injected.

The runtime results should be interpreted separately from the Big-Bench Hard scores. While the Big-Bench Hard scores are reported on Big-Bench Hard, the total runtime is measured separately on 10,000 AG\_News test queries using Llama-3.1-8B on a single NVIDIA A6000 GPU. Under this setting, 10-shot ICL requires longer runtime than LoRA and SITE, mainly due to the inference overhead from repeatedly processing long demonstration-containing prompts. SITE, a recently proposed ICL-derived embedding-based method, shows runtime comparable to LoRA, whereas FV requires substantially longer runtime due to the additional search procedure for selecting intervention heads. For methods requiring task-specific training, the reported runtime includes both training and evaluation time.

Importantly, this analysis should not be interpreted as a general ranking of adaptation paradigms. The reported scores and runtimes are based on a limited set of representative methods, a single backbone model, selected benchmarks, and specific experimental settings. Therefore, they may not fully reflect general trends across different model families, model scales, tasks, implementations, or hardware environments. Rather, this comparison provides a limited empirical snapshot that illustrates the broader qualitative trade-offs discussed in this survey.

\begin{table*}[t]
\centering
\setlength{\tabcolsep}{5pt}
\resizebox{\textwidth}{!}{
\begin{tabular}{llccc}
\toprule
Adaptation Type & Method & \# Trainable Parameters & Big-Bench Hard & Total Runtime \\
\midrule
\midrule
\multirow{2}{*}{Weight-Based Adaptation}
& LoRA~\citep{hu2022lora} & 3407.87K & \textbf{60.39} $\pm$ (0.48) & 171.2 min \\
& (IA)$^3$~\citep{liu2022fewshot} & 524.29K & 60.29 $\pm$ (0.70) & N/A \\
\midrule
\multirow{1}{*}{Prompt-Based Adaptation}
& 10-shot ICL~\citep{brown2020language} & - & 47.17 $\pm$ (0.98) & 248.4 min \\
\midrule
\multirow{4}{*}{Embedding-Based Adaptation~(ICL-derived)}
& FV~\citep{todd2024function} & - & 17.82 $\pm$ (0.37) & 411.2 min \\
& MTV~\citep{huang2024multimodal} & 1.02K & 42.54 $\pm$ (0.71) & 328.8 min \\
& I2CL~\citep{li2025implicit} & 0.13K & 50.60 $\pm$ (1.12) & 178.7 min \\
& SITE~\citep{park2025softinjectiontaskembeddings} & 1.02K & 58.04 $\pm$ (0.72) & 172.0 min \\
\midrule
\multirow{2}{*}{Embedding-Based Adaptation~(Learned)}
& Prompt Tuning~\citep{lester-etal-2021-power} & 81.92K & 33.45 $\pm$ (0.81) & N/A \\
& LIVE~\citep{peng2024live} & 131.10K & 12.89 $\pm$ (2.32) & N/A \\
\bottomrule
\end{tabular}
}
\caption{
Quantitative comparison of representative task adaptation methods, based on the results reported in prior work~\citep{park2025softinjectiontaskembeddings}. 
Big-Bench Hard~\citep{suzgun2023challenging} scores are evaluated using Llama-3.1-8B~\citep{grattafiori2024llama}. 
Total runtime is measured separately on 10,000 AG\_News~\citep{zhang2015character} test queries using Llama-3.1-8B on a single NVIDIA A6000 GPU, and is intended only as an efficiency comparison under this fixed setting. 
For methods requiring task-specific training, runtime includes both training and evaluation time; for FV, it also includes the time of the search procedure for selecting attention heads to intervene on.
}
\label{tab:bbh_grouped_results_params_runtime}
\end{table*}

\subsection{Paper Collection Process}
\label{sec:paper_collection_process}

We describe the paper collection process used to construct the taxonomy in this survey. Our goal is to cover representative and influential work across weight-based, prompt-based, and embedding-based adaptation, rather than to provide an exhaustive list of all related papers.

We first identified foundational papers in each paradigm. We then expanded the candidate set by examining papers that cite these works, prioritizing highly cited papers, papers published in major NLP or machine learning venues, and papers frequently used as baselines in subsequent studies. To improve coverage of recent developments, we also monitored recent-paper platforms such as Trending Papers on Hugging Face~\citep{trending-papers} and screened newly released works relevant to efficient task adaptation.

This process was applied iteratively: after adding new papers, we reviewed their citations, baselines, and related-work discussions to identify additional relevant methods. We repeated this process until the collected set covered the main methodological directions in each paradigm. Since research on LLM adaptation is rapidly evolving, some recent or concurrent works may still be missing. We therefore view the taxonomy as a structured snapshot of the literature at the time of writing.


\begin{table*}[]
\resizebox{\textwidth}{!}{%
\begin{tabular}{cllll}
\toprule
Category &
  Benchmark &
  Output Type &
  Size &
  Description \\
\midrule
\multirow{38}{*}{\begin{tabular}[c]{@{}c@{}}Natural \\ Language \\ Understanding\end{tabular}} &
  \begin{tabular}[c]{@{}l@{}}TREC\\ \citep{li2002learning}\end{tabular} &
  Classification &
  $\sim$6.0K &
  \begin{tabular}[c]{@{}l@{}}TREC Question Classification is a question classification\\ benchmark consisting of open-domain questions annotated with a\\ hierarchical taxonomy of six coarse and fifty fine-grained\\ semantic classes, designed to evaluate question intent\\ understanding and answer type prediction.\end{tabular} \\ \cline{2-5}
 &
  \begin{tabular}[c]{@{}l@{}}CoNLL-2003\\ \citep{sang2003introduction}\end{tabular} &
  Classification &
  $\sim$22K &
  \begin{tabular}[c]{@{}l@{}}CoNLL-2003 is a widely used named entity recognition benchmark \\ focusing on the identification of four entity types including persons, \\ locations, organizations, and miscellaneous names, annotated in \\ English and German newswire text to evaluate information extraction.\end{tabular} \\ \cline{2-5}
 &
  \begin{tabular}[c]{@{}l@{}}Subj\\ \citep{pang2005seeing}\end{tabular} &
  Classification &
  $\sim$10K &
  \begin{tabular}[c]{@{}l@{}}Subjectivity detection dataset consisting of 5,000 subjective\\ movie reviews and 5,000 objective plot summaries designed to test\\ the ability of language models to distinguish between fact-based\\ objective descriptions and opinion-based subjective statements.\end{tabular} \\ \cline{2-5}
 &
  \begin{tabular}[c]{@{}l@{}}SST-5\\ \citep{socher2013recursive}\end{tabular} &
  Classification &
  $\sim$12K &
  \begin{tabular}[c]{@{}l@{}}SST-5 is a fine-grained sentiment analysis dataset providing five\\ sentiment labels ranging from very negative to very positive,\\ designed to evaluate models’ ability to capture subtle emotional\\ shifts and semantic compositionality in movie review phrases.\end{tabular} \\ \cline{2-5}
 &
  \begin{tabular}[c]{@{}l@{}}SNLI\\ \citep{bowman2015large}\end{tabular} &
  Classification &
  $\sim$570K &
  \begin{tabular}[c]{@{}l@{}}SNLI is a large-scale dataset of human-written English sentence\\ pairs manually labeled into entailment, contradiction, and neutral\\ categories, designed to evaluate models’ ability to capture\\ fundamental logical and semantic relationships.\end{tabular} \\ \cline{2-5}
 &
  \begin{tabular}[c]{@{}l@{}}DBPedia\\ \citep{zhang2015character}\end{tabular} &
  Classification &
  $\sim$630K &
  \begin{tabular}[c]{@{}l@{}}DBPedia is a large-scale topic classification benchmark \\ consisting of Wikipedia articles labeled with 14 non-overlapping \\ DBpedia ontology classes, designed to evaluate models’ ability to \\ perform document-level topic and semantic classification.\end{tabular} \\ \cline{2-5}
 &
  \begin{tabular}[c]{@{}l@{}}WikiQA\\ \citep{yang2015wikiqa}\end{tabular} &
  Classification &
  $\sim$3.0K &
  \begin{tabular}[c]{@{}l@{}}WikiQA is an open-domain question answering benchmark consisting\\ of question–sentence pairs derived from Wikipedia search logs,\\ annotated to evaluate answer sentence selection by identifying\\ sentences that contain the correct answer.\end{tabular} \\ \cline{2-5}
 &
  \begin{tabular}[c]{@{}l@{}}GLUE\\ \citep{wang2018glue}\end{tabular} &
  Classification &
  $\sim$1.05M &
  \begin{tabular}[c]{@{}l@{}}GLUE is a unified multi-task benchmark for evaluating general\\ natural language understanding, consisting of nine established NLU\\ tasks: CoLA, SST-2, MRPC, STS-B, QQP, MNLI, QNLI, RTE, and WNLI,\\ covering linguistic acceptability, sentiment analysis, paraphrase\\ detection, semantic similarity, and natural language inference.\end{tabular} \\ \cline{2-5}
 &
  \begin{tabular}[c]{@{}l@{}}SuperGLUE\\ \citep{wang2019superglue}\end{tabular} &
  \begin{tabular}[c]{@{}l@{}}Classification / \\ Multiple Choice Question\end{tabular} &
  $\sim$184K &
  \begin{tabular}[c]{@{}l@{}}SuperGLUE is a more challenging successor to GLUE, designed to \\ evaluate advanced natural language understanding beyond GLUE’s \\ saturation point. It consists of eight established tasks including BoolQ, \\ CB, COPA, MultiRC, ReCoRD, RTE, WiC, and WSC, and it targets \\complex reasoning, commonsense inference, word sense disambiguation, \\and coreference resolution.\end{tabular} \\ \cline{2-5}
 &
  \begin{tabular}[c]{@{}l@{}}ANLI\\ \citep{nie2020adversarial}\end{tabular} &
  Classification &
  $\sim$168K &
  \begin{tabular}[c]{@{}l@{}}ANLI is a challenging NLI benchmark constructed through an\\ iterative, adversarial human-and-model-in-the-loop process over\\ three rounds (R1-R3), targeting examples that strong existing\\ models fail to solve and providing a robust evaluation of advanced\\ natural language inference.\end{tabular} \\
\midrule
\multirow{34}{*}{Reasoning} &
  \begin{tabular}[c]{@{}l@{}}ARC\\ \citep{clark2018think}\end{tabular} &
  Multiple Choice Question &
  $\sim$7.8K &
  \begin{tabular}[c]{@{}l@{}}ARC is a multiple-choice grade-school science QA benchmark \\ split into Easy and Challenge sets, designed to test scientific\\ reasoning beyond surface-level pattern matching.\end{tabular} \\ \cline{2-5}
 &
  \begin{tabular}[c]{@{}l@{}}CommonsenseQA\\ \citep{talmor2019commonsenseqa}\end{tabular} &
  Multiple Choice Question &
  $\sim$12K &
  \begin{tabular}[c]{@{}l@{}}CommonsenseQA is a 5-way multiple-choice commonsense question\\ answering benchmark grounded in the ConceptNet knowledge graph,\\ where questions are authored to discriminate among concept\\ candidates sharing a semantic relation, designed to evaluate\\ diverse commonsense reasoning about everyday situations, physical\\ properties, and social interactions.\end{tabular} \\ \cline{2-5}
 &
  \begin{tabular}[c]{@{}l@{}}HellaSwag\\ \citep{zellers2019hellaswag}\end{tabular} &
  Multiple Choice Question &
  $\sim$60K &
  \begin{tabular}[c]{@{}l@{}}HellaSwag is designed to evaluate commonsense reasoning through \\ sentence-completion tasks. It presents everyday scenarios with four \\ multiple-choice endings and asks models to select the most plausible next \\event; it uses adversarial filtering to generate challenging distractor endings \\that are difficult for current models but typically easy for human evaluators.\end{tabular} \\ \cline{2-5}
 &
  \begin{tabular}[c]{@{}l@{}}MMLU\\ \citep{hendrycks2020measuring}\end{tabular} &
  Multiple Choice Question &
  $\sim$16K &
  \begin{tabular}[c]{@{}l@{}}MMLU is a comprehensive multiple-choice benchmark covering 57\\ subjects across STEM, humanities, and social sciences, designed to\\ evaluate models’ broad world knowledge and knowledge-intensive\\ problem-solving ability across a wide range of difficulty levels.\end{tabular} \\ \cline{2-5}
 &
  \begin{tabular}[c]{@{}l@{}}WinoGrande\\ \citep{sakaguchi2021winogrande}\end{tabular} &
  Multiple Choice Question &
  $\sim$44K &
  \begin{tabular}[c]{@{}l@{}}WinoGrande is a large-scale adversarial commonsense reasoning\\ benchmark of Winograd-style pronoun resolution problems formulated\\ as binary-choice sentence completion, designed to evaluate whether\\ models can use contextual commonsense rather than shallow\\ statistical cues; it applies AFLITE-based debiasing/adversarial\\ filtering to reduce dataset-specific biases.\end{tabular} \\ \cline{2-5}
 &
  \begin{tabular}[c]{@{}l@{}}BBH\\ \citep{suzgun2023challenging}\end{tabular} &
  \begin{tabular}[c]{@{}l@{}}Multiple Choice Question / \\ Open-Ended Generation\end{tabular} &
  $\sim$6.5K &
  \begin{tabular}[c]{@{}l@{}}BBH is a curated subset of BIG-bench~\citep{srivastava2023beyond} consisting of \\challenging reasoning tasks where prior language models underperformed, \\designed to test advanced multi-step reasoning across diverse problem types.\end{tabular} \\ \cline{2-5}
 &
  \begin{tabular}[c]{@{}l@{}}GPQA\\ \citep{rein2024gpqa}\end{tabular} &
  Multiple Choice Question &
  $\sim$0.4K &
  \begin{tabular}[c]{@{}l@{}}GPQA is a graduate-level multiple-choice benchmark written by domain \\experts in biology, physics, and chemistry, designed to evaluate scientific \\reasoning and problem solving with questions intended to be ``Google-proof,'' \\i.e., difficult for skilled non-experts even with unrestricted web access.\end{tabular} \\ \cline{2-5}
 &
  \begin{tabular}[c]{@{}l@{}}MMLU-Pro\\ \citep{wang2024mmlu}\end{tabular} &
  Multiple Choice Question &
  $\sim$12K &
  \begin{tabular}[c]{@{}l@{}}MMLU-Pro is a multi-task multiple-choice benchmark with questions drawn \\ from 14 broad disciplines, constructed by filtering and refining MMLU-style \\ items and adding more reasoning-focused questions from additional sources, \\ and it uses an expanded answer set with ten options per question to better \\ probe challenging understanding and reasoning.\end{tabular} \\
\bottomrule
\end{tabular}%
}
\caption{\textbf{Comprehensive Overview of Benchmarks for LLM Task Adaptation (Part 1 of 4).}}
\label{tab:benchmark (part 1 of 4)}
\end{table*}
\begin{table*}[]
\resizebox{\textwidth}{!}{%
\begin{tabular}{cllll}
\toprule
Category &
  Benchmark &
  Output Type &
  Size &
  Description \\
\midrule
\multirow{28}{*}{Mathematics} &
  \begin{tabular}[c]{@{}l@{}}AQuA-RAT\\ \citep{ling2017program}\end{tabular} &
  \begin{tabular}[c]{@{}l@{}}Multiple Choice Question / \\ Open-Ended Generation\end{tabular} &
  $\sim$100K &
  \begin{tabular}[c]{@{}l@{}}AQuA-RAT is an algebraic word problem 5-way multiple-choice\\ benchmark where each question is paired with a step-by-step\\ natural-language rationale (often including human-readable math\\ expressions), designed to evaluate interpretable multi-step\\ mathematical reasoning beyond predicting only the final answer.\end{tabular} \\ \cline{2-5}
 &
  \begin{tabular}[c]{@{}l@{}}MATH\\ \citep{hendrycks2021measuring}\end{tabular} &
  Open-Ended Generation &
  $\sim$13K &
  \begin{tabular}[c]{@{}l@{}}MATH is a competition-level mathematics problem-solving benchmark\\ covering diverse topics (e.g., algebra, geometry, number theory,\\ counting \& probability, and precalculus), where each problem includes \\ a full step-by-step solution and a final answer, designed to evaluate \\ advanced mathematical reasoning and multi-step problem solving.\end{tabular} \\ \cline{2-5}
 &
  \begin{tabular}[c]{@{}l@{}}SVAMP\\ \citep{patel2021nlp}\end{tabular} &
  Open-Ended Generation &
  $\sim$1.0K &
  \begin{tabular}[c]{@{}l@{}}SVAMP is an adversarial challenge benchmark of one-unknown\\ arithmetic math word problems created by applying small but\\ systematic variations to existing problems, designed to evaluate\\ robust multi-step arithmetic reasoning beyond keyword matching and\\ shallow statistical cues.\end{tabular} \\ \cline{2-5}
 &
  \begin{tabular}[c]{@{}l@{}}GSM8K\\ \citep{cobbe2021training}\end{tabular} &
  Open-Ended Generation &
  $\sim$8.8K &
  \begin{tabular}[c]{@{}l@{}}GSM8K is a grade-school math word problem benchmark consisting \\ of natural-language questions with final numeric answers, designed to \\evaluate multi-step arithmetic reasoning required to solve the problems.\end{tabular} \\ \cline{2-5}
 &
  \begin{tabular}[c]{@{}l@{}}TabMWP\\ \citep{lu2022dynamic}\end{tabular} &
  \begin{tabular}[c]{@{}l@{}}Multiple Choice Question / \\ Open-Ended Generation\end{tabular} &
  $\sim$38K &
  \begin{tabular}[c]{@{}l@{}}TabMWP is a tabular math word problem benchmark where each\\ question is paired with a tabular context provided in multiple\\ formats (e.g., table image and structured text), designed to\\ evaluate joint reasoning over tables and natural-language\\ descriptions for deriving correct numerical answers (with gold\\ step-by-step solutions available).\end{tabular} \\ \cline{2-5}
 &
  \begin{tabular}[c]{@{}l@{}}MATH500\\ \citep{lightman2023let}\end{tabular} &
  Open-Ended Generation &
  $\sim$0.5K &
  \begin{tabular}[c]{@{}l@{}}MATH500 is a curated evaluation subset of the MATH dataset,\\ designed to provide an efficient yet diverse test of\\ competition-style mathematical problem solving.\end{tabular} \\ \cline{2-5}
 &
  \begin{tabular}[c]{@{}l@{}}MetaMathQA\\ \citep{yu2023metamath}\end{tabular} &
  Open-Ended Generation &
  $\sim$395K &
  \begin{tabular}[c]{@{}l@{}}MetaMathQA is an augmented mathematical reasoning dataset\\ bootstrapped from the training sets of GSM8K and MATH by generating \\ diverse, semantically equivalent problem variants (with answer \\ augmentation), designed to improve models' robustness in multi-step \\ mathematical reasoning across varied linguistic formulations.\end{tabular} \\
\midrule
\multirow{20}{*}{Coding} &
  \begin{tabular}[c]{@{}l@{}}HumanEval\\ \citep{chen2021evaluating}\end{tabular} &
  Open-Ended Generation &
  $\sim$0.2K &
  \begin{tabular}[c]{@{}l@{}}HumanEval is a benchmark of handwritten Python programming \\ tasks specified by function signatures and docstrings, designed to\\ evaluate models' ability to generate correct code solutions for\\ the given problem specifications.\end{tabular} \\ \cline{2-5}
 &
  \begin{tabular}[c]{@{}l@{}}MBPP\\ \citep{austin2021program}\end{tabular} &
  Open-Ended Generation &
  $\sim$1.0K &
  \begin{tabular}[c]{@{}l@{}}MBPP is a benchmark of entry-level Python programming tasks\\ specified by short natural-language problem descriptions, designed\\ to evaluate models' ability to synthesize correct short programs\\ involving basic algorithms and common data-structure manipulations.\end{tabular} \\ \cline{2-5}
 &
  \begin{tabular}[c]{@{}l@{}}MultiPL-E\\ \citep{cassano2023multipl}\end{tabular} &
  Open-Ended Generation &
  $\sim$25K &
  \begin{tabular}[c]{@{}l@{}}MultiPL-E is a polyglot code generation benchmark that translates\\ Python-based programming tasks from HumanEval and MBPP into many\\ target languages, designed to evaluate models' cross-language code\\ synthesis ability and consistency across diverse programming languages.\end{tabular} \\ \cline{2-5}
 &
  \begin{tabular}[c]{@{}l@{}}SWE-bench\\ \citep{jimenez2023swe}\end{tabular} &
  Open-Ended Generation &
  $\sim$2.3K &
  \begin{tabular}[c]{@{}l@{}}SWE-bench is a software engineering benchmark built from real\\ GitHub issues and corresponding pull requests across multiple\\ popular repositories, designed to evaluate models' ability to\\ understand and modify large codebases by producing code changes\\ that resolve the described issues (e.g., bug fixes or feature requests).\end{tabular} \\ \cline{2-5}
 &
  \begin{tabular}[c]{@{}l@{}}CruxEval\\ \citep{gu2024cruxeval}\end{tabular} &
  Open-Ended Generation &
  $\sim$0.8K &
  \begin{tabular}[c]{@{}l@{}}CruxEval is a Python function reasoning benchmark consisting of\\ short functions paired with input-output examples, designed to\\ evaluate models' ability to understand program execution by\\ performing output prediction (infer the output for a given input)\\ and input prediction (find an input that produces a given output).\end{tabular} \\
\bottomrule
\end{tabular}%
}
\caption{\textbf{Comprehensive Overview of Benchmarks for LLM Task Adaptation (Part 2 of 4).}}
\label{tab:benchmark (part 2 of 4)}
\end{table*}
\begin{table*}[]
\resizebox{\textwidth}{!}{%
\begin{tabular}{cllll}
\toprule
Category &
  Benchmark &
  Output Type &
  Size &
  Description \\
\midrule
\multirow{14}{*}{\begin{tabular}[c]{@{}c@{}}Question \\ Answering\end{tabular}} &
  \begin{tabular}[c]{@{}l@{}}SQuAD v1.1\\ \citep{rajpurkar2016squad}\end{tabular} &
  Open-Ended Generation &
  $\sim$98K &
  \begin{tabular}[c]{@{}l@{}}SQuAD v1.1 is an extractive reading comprehension benchmark built\\ from questions written on Wikipedia passages, where each question\\ is paired with an answer that is a contiguous text span from the\\ given passage, designed to evaluate models’ ability to perform\\ span-based question answering.\end{tabular} \\ \cline{2-5}
 &
  \begin{tabular}[c]{@{}l@{}}TriviaQA\\ \citep{joshi2017triviaqa}\end{tabular} &
  Open-Ended Generation &
  $\sim$96K &
  \begin{tabular}[c]{@{}l@{}}TriviaQA is a reading-comprehension benchmark built from trivia\\ questions paired with evidence documents from Wikipedia and the web, \\ designed to test answering complex, compositional questions despite \\ substantial mismatch between questions and supporting evidence.\end{tabular} \\ \cline{2-5}
 &
  \begin{tabular}[c]{@{}l@{}}HotpotQA\\ \citep{yang2018hotpotqa}\end{tabular} &
  Open-Ended Generation &
  $\sim$113K &
  \begin{tabular}[c]{@{}l@{}}HotpotQA is a multi-hop QA benchmark where each question is paired\\ with supporting facts across multiple Wikipedia articles, designed\\ to test whether models can integrate evidence from more than one\\ document to answer complex questions.\end{tabular} \\ \cline{2-5}
 &
  \begin{tabular}[c]{@{}l@{}}Natural Questions\\ \citep{kwiatkowski2019natural}\end{tabular} &
  Open-Ended Generation &
  $\sim$323K &
  \begin{tabular}[c]{@{}l@{}}Natural Questions is an open-domain QA benchmark built from real\\ Google search queries paired with Wikipedia pages, providing\\ annotations for both long answers (passages) and short answers\\ (specific entities) to test end-to-end question answering grounded\\ in retrieved evidence.\end{tabular} \\
\midrule
\multirow{20}{*}{Summarization} &
  \begin{tabular}[c]{@{}l@{}}CNN/DailyMail\\ \citep{hermann2015teaching}\end{tabular} &
  Open-Ended Generation &
  $\sim$312K &
  \begin{tabular}[c]{@{}l@{}}CNN/DailyMail is a news summarization benchmark consisting of full\\ news articles paired with human-written highlights (bullet-style\\ summary sentences), designed to evaluate models’ ability to\\ produce concise summaries that capture the key information in\\ long-form news reports.\end{tabular} \\ \cline{2-5}
 &
  \begin{tabular}[c]{@{}l@{}}XSum\\ \citep{narayan2018don}\end{tabular} &
  Open-Ended Generation &
  $\sim$227K &
  \begin{tabular}[c]{@{}l@{}}XSum is an extreme abstractive summarization benchmark pairing BBC\\ news articles with single-sentence summaries that capture what the\\ article is about, designed to measure models’ ability to produce\\ highly condensed, gist-focused summaries rather than\\ detail-preserving paraphrases.\end{tabular} \\ \cline{2-5}
 &
  \begin{tabular}[c]{@{}l@{}}SAMSum\\ \citep{gliwa2019samsum}\end{tabular} &
  Open-Ended Generation &
  $\sim$16K &
  \begin{tabular}[c]{@{}l@{}}SAMSum is an abstractive dialogue summarization benchmark\\ consisting of messenger-style chat conversations with\\ human-written summaries, designed to evaluate models’ ability to\\ summarize informal, multi-speaker dialogues that may include\\ colloquial language such as slang and emoticons.\end{tabular} \\ \cline{2-5}
 &
  \begin{tabular}[c]{@{}l@{}}DialogSum\\ \citep{chen2021dialogsum}\end{tabular} &
  Open-Ended Generation &
  $\sim$13K &
  \begin{tabular}[c]{@{}l@{}}DialogSum is an abstractive dialogue summarization benchmark\\ consisting of multi-turn dialogues from diverse real-life\\ scenarios (compiled from multiple dialogue sources) paired with\\ human-written summaries and topics, designed to evaluate models'\\ ability to capture salient information and speaker intents in\\ multi-speaker conversations\end{tabular} \\ \cline{2-5}
 &
  \begin{tabular}[c]{@{}l@{}}XL-Sum\\ \citep{hasan2021xl}\end{tabular} &
  Open-Ended Generation &
  $\sim$1.4M &
  \begin{tabular}[c]{@{}l@{}}XL-Sum is a large-scale multilingual abstractive summarization\\ benchmark consisting of BBC news article–summary pairs across\\ dozens of languages, designed to evaluate summarization capability\\ across both high- and low-resource languages.\end{tabular} \\
\midrule
\multirow{10}{*}{\begin{tabular}[c]{@{}c@{}}Structured \\ Data-to-Text\end{tabular}} &
  \begin{tabular}[c]{@{}l@{}}E2E NLG\\ \citep{novikova2017e2e}\end{tabular} &
  Open-Ended Generation &
  $\sim$51K &
  \begin{tabular}[c]{@{}l@{}}E2E NLG is a restaurant-domain data-to-text generation benchmark that pairs \\ dialogue-act-style meaning representations in the form of attribute–value pairs \\ with human-written utterances, and is designed to evaluate end-to-end \\ verbalization and content selection under diverse surface realizations.\end{tabular} \\ \cline{2-5}
 &
  \begin{tabular}[c]{@{}l@{}}WebNLG\\ \citep{gardent2017webnlg}\end{tabular} &
  Open-Ended Generation &
  $\sim$22K &
  \begin{tabular}[c]{@{}l@{}}WebNLG is a data-to-text benchmark that maps sets of DBpedia RDF triples \\ to short natural-language texts, and is designed to evaluate micro-planning \\ in generation, including lexicalization, aggregation, referring expression \\ generation, and sentence segmentation, while preserving the input facts.\end{tabular} \\ \cline{2-5}
 &
  \begin{tabular}[c]{@{}l@{}}ToTTo\\ \citep{parikh2020totto}\end{tabular} &
  Open-Ended Generation &
  $\sim$136K &
  \begin{tabular}[c]{@{}l@{}}ToTTo is a controlled table-to-text generation benchmark pairing\\ Wikipedia tables with a set of highlighted cells and a\\ human-written one-sentence description, designed to evaluate\\ grounded and faithful text generation from structured tabular data\\ under explicit content selection.\end{tabular} \\
\bottomrule
\end{tabular}%
}
\caption{\textbf{Comprehensive Overview of Benchmarks for LLM Task Adaptation (Part 3 of 4).}}
\label{tab:benchmark (part 3 of 4)}
\end{table*}
\begin{table*}[]
\resizebox{\textwidth}{!}{%
\begin{tabular}{cllll}
\toprule
Category &
  Benchmark &
  Output Type &
  Size &
  Description \\
\midrule
\multirow{42}{*}{\begin{tabular}[c]{@{}c@{}}Safety and \\ Trustworthiness\end{tabular}} &
  \begin{tabular}[c]{@{}l@{}}HateSpeech18\\ \citep{de2018hate}\end{tabular} &
  Classification &
  $\sim$11K &
  \begin{tabular}[c]{@{}l@{}}HateSpeech18 is a sentence-level hate speech dataset sampled from\\ posts on the Stormfront white-supremacist forum and manually\\ labeled as hate vs non-hate, designed to evaluate models’ ability\\ to detect hate speech in highly domain-specific, ideologically\\ skewed online discussions.\end{tabular} \\ \cline{2-5}
 &
  \begin{tabular}[c]{@{}l@{}}RealToxicityPrompts\\ \citep{gehman2020realtoxicityprompts}\end{tabular} &
  Open-Ended Generation &
  $\sim$100K &
  \begin{tabular}[c]{@{}l@{}}RealToxicityPrompts is a dataset of 100k naturally occurring,\\ sentence-level prompts drawn from English web text and paired with\\ toxicity annotations, designed to evaluate whether language models\\ produce toxic continuations when conditioned on real-world prompts\\ spanning a range of toxicity levels.\end{tabular} \\ \cline{2-5}
 &
  \begin{tabular}[c]{@{}l@{}}CrowS-Pairs\\ \citep{nangia2020crows}\end{tabular} &
  Multiple Choice Question &
  $\sim$1.5K &
  \begin{tabular}[c]{@{}l@{}}CrowS-Pairs is a social bias evaluation benchmark consisting of\\ sentence pairs that differ in whether they express a stereotype,\\ covering nine bias categories (e.g., race, gender, religion, age),\\ designed to measure models’ tendency to prefer stereotypical\\ statements over less-stereotyping alternatives.\end{tabular} \\ \cline{2-5}
 &
  \begin{tabular}[c]{@{}l@{}}ToxiGen\\ \citep{hartvigsen2022toxigen}\end{tabular} &
  Classification &
  $\sim$274K &
  \begin{tabular}[c]{@{}l@{}}ToxiGen is a large-scale machine-generated toxicity dataset\\ consisting of toxic and benign statements about minority identity\\ groups, created to surface implicit and adversarially crafted\\ toxic language beyond explicit slurs or profanity, and designed to\\ evaluate models’ ability to detect subtle harmful content rather\\ than relying on group-mention shortcuts.\end{tabular} \\ \cline{2-5}
 &
  \begin{tabular}[c]{@{}l@{}}ParaDetox\\ \citep{logacheva2022paradetox}\end{tabular} &
  Open-Ended Generation &
  $\sim$20K &
  \begin{tabular}[c]{@{}l@{}}ParaDetox is a parallel text detoxification dataset pairing toxic\\ sentences with human-written non-toxic paraphrases, designed to\\ test whether models can remove toxicity while preserving the\\ original meaning.\end{tabular} \\ \cline{2-5}
 &
  \begin{tabular}[c]{@{}l@{}}TruthfulQA\\ \citep{lin2022truthfulqa}\end{tabular} &
  \begin{tabular}[c]{@{}l@{}}Multiple Choice Question / \\ Open-Ended Generation\end{tabular} &
  $\sim$0.8K &
  \begin{tabular}[c]{@{}l@{}}TruthfulQA is a question answering benchmark consisting of\\ questions spanning 38 categories that are crafted to trigger\\ common misconceptions, designed to evaluate whether models can\\ produce truthful, misconception-resistant answers rather than\\ imitating popular human falsehoods.\end{tabular} \\ \cline{2-5}
 &
  \begin{tabular}[c]{@{}l@{}}ETHOS\\ \citep{mollas2022ethos}\end{tabular} &
  Classification &
  $\sim$1.0K &
  \begin{tabular}[c]{@{}l@{}}ETHOS is an online hate-speech detection dataset of YouTube and\\ Reddit comments with both a binary hate-speech label and an\\ additional multi-label variant that annotates hateful comments\\ with fine-grained hate categories.\end{tabular} \\ \cline{2-5}
 &
  \begin{tabular}[c]{@{}l@{}}HaluEval\\ \citep{li2023halueval}\end{tabular} &
  Classification &
  $\sim$35K &
  \begin{tabular}[c]{@{}l@{}}HaluEval is a hallucination benchmark built from instruction-style\\ prompts and task-specific contexts (QA, dialogue, and\\ summarization), pairing responses with annotations of whether\\ content is hallucinated to assess factuality and faithfulness in\\ generated text.\end{tabular} \\ \cline{2-5}
 &
  \begin{tabular}[c]{@{}l@{}}Do-Not-Answer\\ \citep{wang2023not}\end{tabular} &
  Open-Ended Generation &
  $\sim$0.9K &
  \begin{tabular}[c]{@{}l@{}}Do-Not-Answer is a safety safeguard evaluation benchmark\\ consisting of risky questions and instructions spanning a\\ structured taxonomy of harm types (e.g., privacy leakage, illegal\\ activities, toxic content, misinformation harms, and human–chatbot\\ interaction risks), designed to assess whether models can avoid\\ facilitating harm and respond responsibly to unsafe requests.\end{tabular} \\
\midrule
\multirow{5}{*}{\begin{tabular}[c]{@{}c@{}}General ICL \\ Capability\end{tabular}} &
  \begin{tabular}[c]{@{}l@{}}FV Benchmark\\ \citep{todd2024function}\end{tabular} &
  \begin{tabular}[c]{@{}l@{}}Multiple Choice Question / \\ Open-Ended Generation\end{tabular} &
  $\sim$9.3K &
  \begin{tabular}[c]{@{}l@{}}FV Benchmark comprises 29 abstractive and 28 extractive tasks, \\including closely related and contrasting variants, making it \\well suited for studying in-context learning behaviors. \\Abstractive tasks require generating information not explicitly \\present in the prompt, whereas extractive tasks involve directly \\retrieving the answer from it.\end{tabular} \\ \cline{2-5}
 &
  \begin{tabular}[c]{@{}l@{}}ICL-50\\ \citep{zhang2025more}\end{tabular} &
  \begin{tabular}[c]{@{}l@{}}Multiple Choice Question / \\ Open-Ended Generation\end{tabular} &
  $\sim$3.2M&
  \begin{tabular}[c]{@{}l@{}}ICL-50 is a many-shot in-context learning dataset spanning 50\\ tasks across several task families, created to study how model\\ performance and behavior change as the number of in-context\\ examples scales to long contexts.\end{tabular} \\
  \bottomrule
\end{tabular}%
}
\caption{\textbf{Comprehensive Overview of Benchmarks for LLM Task Adaptation (Part 4 of 4).}}
\label{tab:benchmark (part 4 of 4)}
\end{table*}


\section{Use of LLMs in This Work} 
We used chat-based LLMs for sentence-level editing to check grammar and improve clarity during paper writing. All edits were reviewed and verified by the authors. All conceptual contributions are solely by the authors.

\end{document}